\documentclass{article}

\usepackage{arxiv}

\usepackage[utf8]{inputenc} 
\usepackage[T1]{fontenc}    
\usepackage{hyperref}       
\usepackage{url}            
\usepackage{booktabs}       
\usepackage{amsmath, amssymb, amsfonts}
\usepackage{nicefrac}       
\usepackage{microtype}      
\usepackage{graphicx}       
\usepackage{multirow}       
\usepackage{threeparttable} 
\usepackage{caption}
\usepackage{subcaption}
\usepackage{longtable}
\usepackage{tabularx}
\usepackage{rotating}
\usepackage[table]{xcolor}
\usepackage{adjustbox}
\usepackage{float}
\usepackage{natbib}
\usepackage{doi}
\usepackage{makecell}
\usepackage{indentfirst}
\usepackage{enumitem}
\usepackage{comment}
\usepackage{booktabs}

\title{Populism Meets AI: \\Advancing Populism Research with LLMs}
\author{
\small
  Yujin J. Jung\thanks{Corresponding author}\\
  Mount St. Mary’s University\\
  \texttt{yujinjuliajung@gmail.com}
  \And
  Eduardo Ryô Tamaki\\
  German Institute for Global and Area Studies\\
  \texttt{eduardo.tamaki@giga-hamburg.de}
  \AND
  Julia Chatterley\\
  Princeton University\\
  \texttt{jc3907@princeton.edu}
  \And
  Grant Mitchell\\
  University of California, Los Angeles\\
  \texttt{gamitchell@ucla.edu}
  \AND
  Semir Dzebo\\
  University of Oxford\\
  \texttt{semir.dzebo@csls.ox.ac.uk}
  \And
  Cristóbal Sandoval\\
  Diego Portales University\\
  \texttt{cristobal.sandoval@mail.udp.cl}
  \AND
  Levente Littvay\\
  ELTE Centre for Social Sciences\\
  \texttt{littvay.levente@tk.elte.hu}
  \And
  Kirk A. Hawkins\\
  Brigham Young University\\
  \texttt{kirk\_hawkins@byu.edu}
}

\date{} 

\renewcommand{\shorttitle}{Populism Meets AI: Advancing Populism Research with LLMs}

\hypersetup{
  pdftitle={Populism Meets AI: Advancing Populism Research with LLMs},
  pdfsubject={Computational Social Science, Political Text Analysis},
  pdfauthor={Eduardo Ryô Tamaki, Yujin J. Jung, Julia Chatterley, Grant Mitchell, Semir Dzebo, Cristóbal Sandoval, Levente Littvay, Kirk A. Hawkins},
  pdfkeywords={Populism, Large Language Models, Holistic Grading, Computational Political Science, Text Analysis},
  colorlinks=true,
  linkcolor=blue,
  citecolor=blue,
  urlcolor=blue
}

\graphicspath{{./Figures/}}

\begin{document}
\maketitle

\begin{abstract}
	Measuring the ideational content of populism remains a challenge. Traditional strategies based on textual analysis have been critical for building the field’s foundations and providing a valid, objective indicator of populist framing. Yet these approaches are costly, time-consuming, and difficult to scale across languages, contexts, and large corpora. Here we present the results from a rubric- and anchor-guided chain-of-thought (CoT) prompting approach that mirrors human-coder training. By leveraging the Global Populism Database (GPD), a comprehensive dataset of global leaders' speeches annotated for degrees of populism, we replicate the process used to train human coders by prompting the LLM with an adapted version of the same documentation to guide the model's reasoning. We then test multiple proprietary and open-weight models the LLM by replicating scores in the GPD. Our findings reveal that this domain-specific prompting strategy enables the LLM to achieve classification accuracy on par with expert human coders, demonstrating its ability to navigate the nuanced, context-sensitive aspects of populism.
\end{abstract}



\section{Introduction} 

The global expansion of populist discourse has intensified scholarly attention to the conceptualization and measurement of populism. While important differences remain among competing approaches, most scholars now agree that a defining feature of populism is its discursive construction of politics as a moral struggle between a virtuous common people and a corrupt, conspiring elite. This premise is evident across several leading traditions. The ideational approach defines populism minimally in terms of this people-versus-elite frame (Hawkins et al., 2019; Mudde, 2017). The discourse-theoretic tradition, associated with the Essex school, initially emphasized antagonistic or Manichaean discourse broadly, but has increasingly specified populism as a discourse built around the construction of “people” and “elite” as its constitutive identities (Stavrakakis, 2024). The political-strategic perspective, for its part, prioritizes organizational aspects such as charismatic leadership, but nonetheless treats people-versus-elite rhetoric as one of populism’s distinctive features (Weyland, 2024).
\nocite{hawkins2019ideational,mudde2017ideational,stavrakakis2024populist,weyland2024democracy}

Despite this broad convergence, the measurement of populism continues to pose significant challenges. The principal one is how to measure a set of ideas that is both latent and context-dependent. Unlike programmatic ideologies, populist ideas do not articulate a stable set of policies, but operate as a discourse or frame juxtaposing a morally virtuous “people” against a corrupt and self-serving “elite” (Aslanidis, 2016; Mudde, 2017). Because populist ideas are embedded in rhetorical strategies and can attach themselves to diverse, ideological positions, their measurement requires methods capable of detecting subtle discursive cues rather than fixed vocabularies. 
\nocite{aslanidis2016ideology,mudde2017ideational}

Existing approaches to this task have relied heavily on text-based strategies, either through qualitative coding by trained human coders or through automated machine learning classifiers trained on human annotations. These strategies remain foundational because they draw directly on the words of political actors, providing face-valid and context-sensitive indicators of populist framing. Yet they are also resource-intensive. Human coding, even when carefully trained and calibrated, demands multilingual expertise, extensive coder training, and substantial financial support. Consequently, most projects achieve only limited cross-national and temporal scope, rarely covering more than a few countries or languages simultaneously (Armony \& Armony, 2005; Dai \& Kustov, 2022; Hawkins \& Castanho Silva, 2019; Jagers \& Walgrave, 2007; Rooduijn \& Pauwels, 2011). Even the most ambitious initiative to date—the Global Populism Database, which uses Holistic Grading (HG) to measure populism in the speeches of chief executives across dozens of languages and regions—extends across only two decades (Hawkins et al., 2022).
\nocite{armony2005indictments,dai2022campaign,hawkins2019textual,jagers2007communication,rooduijn2011measuring,hawkins2022global}

Recent advances in large language models (LLMs), more colloquially known as AI, offer a potential solution to these challenges. Initially developed for natural language processing in computer science, LLMs are now increasingly employed in the social sciences to generate synthetic data, conduct interviews, simulate experimental participants, and, most importantly for this study, automate content analysis (Argyle et al., 2023; Tamaki \& Littvay, 2024; Garvey \& Blanchard, 2025). Emerging evidence suggests that LLMs can classify political texts with accuracy approaching that of human coders, whether by labeling topics, detecting sentiment, or scoring ideological orientations (Gilardi, Alizadeh \& Kubli, 2023; Ziems et al., 2024). Their multilingual capabilities further allow them to process diverse corpora without the need for separate native-speaking coders. By prompting LLMs with the same rubrics used in HG, it becomes possible to test whether these systems can approximate the integrated judgments of human graders while overcoming the practical barriers that constrain human-based approaches.
\nocite{argyle2023simulate,tamaki2024chrono,garvey2025lucid,gilardi2023GPT,ziems2024transform}

This paper evaluates whether LLMs can replicate Holistic Grading of populism. We focus on a set of 12 political speeches drawn from the United Kingdom, Turkey, and Montenegro, spanning four speech types (campaign, famous, international, and ribbon-cutting). We compare the performance of multiple LLMs against human-coded benchmarks, with five independent runs per model. We find that structuring LLMs with Holistic Grading works: the strongest models approximate human coders with a high degree of fidelity. However, caveats remain; most systems compress the populism scale, nudging upward scores that should be near zero and pulling downward those that should be maximally populist, though they preserve rank-order consistency across speech categories. Moreover, performance varies across models: GPT-5 (reasoning) achieves the highest consistency and agreement with human coders, while Qwen3 (reasoning) performs similarly, outperforming other open-source systems by a substantial margin.

This study has several important implications. First, it contributes to the broader literature on the measurement of populism by introducing an AI-based methodology that aligns with the conceptual richness of political science. Unlike generic AI models, our approach bridges the gap between computational efficiency and the nuanced understanding required for studying political phenomena. Second, our research highlights the potential for AI to handle complex political science concepts, suggesting that as AI continues to evolve, it may increasingly be used to explore other intricate constructs within the discipline, such as nationalist and pluralist framing. Finally, our project lays the groundwork for the development of publicly accessible AI models capable of classifying populist speeches, paving the way for future research and practical applications in political science and beyond.

The remainder of the paper proceeds in four steps. We first review advances in LLM research relevant to social science applications, with emphasis on content analysis. We then outline our adaptation of Holistic Grading to the LLM context, including the use of chain-of-thought prompting to guide model reasoning. We next present our empirical evaluation, quantifying human–AI agreement through multiple complementary metrics. Finally, we assess error patterns, calibration, and anchor fidelity to evaluate the conditions under which LLMs can function as reliable holistic graders. In doing so, we aim to demonstrate both the promise and the limitations of deploying advanced language models for the systematic measurement of populism.

\section{Measuring Populism} 

A central challenge in the study of populism is how to capture a concept that is both latent and context-dependent. Populism is not reducible to a single issue position or a programmatic ideology, but functions as a master frame through which the sovereign is identified and political conflict is interpreted (Aslanidis, 2016; Jenne, Hawkins \& Castanho Silva, 2021). This framing logic can attach itself to a wide array of ideological projects (Akkerman, Zaslove \& Spruyt, 2017; Rooduijn \& Akkerman, 2017), and is not articulated through canonical texts or codified doctrines in the manner of traditional ideologies. Instead, it often emerges organically from grassroots politics and is activated through familiar tropes that resonate with audiences (Stavrakakis, 2024). The task of measuring populism, therefore, requires methods that can capture latent rhetorical cues, rather than relying on fixed vocabularies or explicit doctrinal statements.
\nocite{aslanidis2016ideology,jenne2021mapping,akkerman2017we,rooduijn2017flank,stavrakakis2024populist}

Unsurprisingly, textual analysis has been foundational in the field. Early works promoting a focus on populist ideas were often case studies using qualitative textual analysis to clarify the concept of populism while providing evidence to categorize parties and politicians as populist (Canovan, 1981; de la Torre, 2000; Hawkins, 2003; Laclau, 1977; Mudde, 2004). Quickly, however, scholars moved to quantitative techniques to provide larger datasets (Armony
\& Armony, 2005; Jagers \& Walgrave, 2007). Often, these were dictionary-based approaches relying on hybrid forms of automated and human-based coding, and they allowed scholars to look intensively at multiple parties and politicians, sometimes across long spans of time (Rooduijn \& Pauwels, 2011; Bonikowski \& Gidron, 2016). However, because dictionaries are expensive to build, these efforts were necessarily limited to small numbers of countries with one or a few languages, and without cross-regional comparison. More recent efforts have shifted to machine-learning approaches, which have the computer generate dictionaries based on pre-coded reference texts, and are in theory cheaper and faster to operate. Yet these too remain limited by the availability of pre-coded datasets and are often confined to narrow genres of text or specific countries (Dai \& Kustov, 2022; Hawkins \& Castanho Silva, 2019). All of these techniques also raise concerns about whether diffuse ideas can be accurately measured with techniques using individual words as the level of analysis (Rooduijn \& Pauwels, 2011).\footnote{One way around this problem is to circumvent textual analysis entirely and rely on expert surveys. These dispense with direct analysis of ideas and instead solicit the judgments of country experts presumed to be familiar with their parties’ rhetoric. Early applications of this method were hampered by reliability concerns, often reflecting the views of small numbers of scholars or even single authors. More recent efforts have drawn on larger pools of experts and more sophisticated survey designs, thereby increasing coverage and comparability (Lindberg et al., 2022; Norris, 2020; Wiesehomeier, Singer \& Ruth-Lovell, 2021; Zaslove, Huber \& Meijers, 2025). Expert surveys also provide valuable ancillary data on party ideology and organization, allowing for broader analyses of party systems. Nevertheless, these surveys face well-known limitations: experts may lack a global frame of reference, leading them to rate parties only relative to domestic competitors, and their judgments are often colored by strong prior views about which actors in their country “count” as populist (Meijers
\& Wiesehomeier, 2023).}
\nocite{canovan1981populism,torre2000seduction,hawkins2003venezuela,laclau1977politics,mudde2004zeitgeist,armony2005indictments,jagers2007communication,rooduijn2011measuring,bonikowski2016style,dai2022campaign,hawkins2019textual,rooduijn2011measuring}
\nocite{lindberg2022vparty,norris2020measuring,wiesehomeier2021prepps,zaslove2025state,meijers2023expert}

Holistic Grading has emerged as a particularly refined approach to these challenges. Unlike dictionary-based methods, which rely on pre-specified word lists, and sentence-by-sentence coding, HG requires coders to read entire texts and assign an integrated score based on the presence and intensity of people-centrism, anti-elitism, and Manichaean framing. By embedding judgment in the full rhetorical and contextual fabric of a speech, HG captures populism’s latent and context-dependent qualities more effectively than surface-level methods. 

An important advantage of HG lies in its ability to produce more valid and reliable measures of populist discourse than competing methods. Because coders are trained to consider whole texts in light of a detailed rubric, HG captures rhetorical strategies that may otherwise be fragmented or inconsistent across smaller units of analysis. This holistic orientation makes it possible to identify patterns that reflect the overall communicative style of political leaders rather than isolated word choices or phrases. In practice, HG has generated data that align more closely with expert judgments and theoretical expectations, thereby reinforcing its reputation as the benchmark method for measurement in populism research (Hawkins et al., 2019; Hawkins et al., 2022; Meijers \& Zaslove, 2021).
\nocite{hawkins2019ideational,hawkins2022global,meijers2021measuring}

Despite its strengths, HG comes with significant costs. Human coders must undergo rigorous training to internalize the coding scheme and to calibrate their judgments to those of other coders. These coders must also meet demanding requirements: they must be fluent in the language of the original speeches, possess a knowledge of politics that allows them to identify subtle rhetorical features, participate in English-language training and calibration sessions, and have sufficient time to code lengthy political speeches. The result is a process that is resource-intensive, both financially and in terms of personnel. Even with coordinated international research teams, sustaining HG projects at scale proves difficult. As a result, while HG remains the “gold standard” for measuring populism, its reliance on scarce human labor limits its scalability across languages, countries, and time periods.

These limitations have prompted growing interest in alternative solutions, particularly those that can replicate the strengths of HG while reducing its costs. Recent advances in large language model research present a promising opportunity to automate elements of this process. By leveraging computational tools capable of identifying latent meanings, scholars may be able to replicate the interpretive depth of HG without relying on teams of extensively trained coders. This possibility motivates the next step of our inquiry: assessing how LLMs can be deployed to measure populism systematically.

\section{Advancements in LLM Research} 

As discussed earlier, HG is particularly well suited for studying populism because it captures subtle and context-dependent meanings that elude conventional text analysis. The emergence of large language models has not only transformed natural language processing but has also revolutionized how social scientists generate and analyze data, offering new possibilities for replicating HG computationally.

Traditional text analysis has relied on surface-level features such as word frequencies, keyword counts, and fixed-label classifications to infer meaning from text. Large language models now provide a more advanced alternative, functioning as semantic reasoners that integrate linguistic relationships within broader contexts and capture latent discursive meaning (Naveed et al., 2025; Mu et al., 2024). Since the introduction of the Transformer architecture (Vaswani et al., 2017), LLMs have evolved from predictive systems into models capable of contextual understanding and causal reasoning, replacing earlier statistical approaches by identifying intent and discourse-level coherence.
\nocite{VaswaniEtAl2017} \nocite{naveed2025comprehensive} \nocite{mu2024large}

Building on this transformation, recent innovations have strengthened LLMs’ capacity for both reasoning and scalability, enabling them to process context-dependent meanings with greater depth and efficiency. The Chain-of-Thought paradigm trains models to articulate intermediate reasoning steps that mirror human logical thinking (Wei et al., 2022), allowing them to explain how conclusions are reached rather than simply generating outputs. This advancement enhances their ability to interpret complex political concepts such as populism or elite discourse, which depend on implicit meanings and contextual variation rather than explicit lexical cues. Complementing this development, the Mixture-of-Experts architecture improves scalability and computational efficiency by selectively activating specialized expert modules based on input characteristics (Riquelme et al., 2021), enabling larger and more adaptive models without increasing computational cost.
\nocite{wei2022chain} \nocite{RiquelmeEtAl2021}

LLMs’ multilingual pretraining capabilities have further extended these advances beyond conventional text-analysis models (Zhao et al., 2024). By training on parallel corpora across hundreds of languages, they share a unified semantic space that allows the consistent capture of equivalent conceptual meanings across diverse linguistic and cultural contexts. Such multilingual architectures provide a crucial advantage for comparative political research, particularly for studying populism, where cross-cultural variation in rhetoric and meaning is central to theory and measurement.
\nocite{zhaoetal2024large}

In addition, the growing ecosystem of open-source models, such as DeepSeek (Guo et al., 2025), Qwen (Bai et al., 2023), and the recently released GPT-OSS (Agarwal et al., 2025), has significantly enhanced accessibility and reproducibility in academic research (Manchanda et al., 2025). Scholars can now directly access model weights, fine-tune models, and design prompts independently, enabling the adaptation of models to local political concepts and discourse contexts. This shift promotes interpretive transparency and supports the cumulative development of computational methods in political science.
\nocite{ManchandaEtAl2024} \nocite{guo2025deepseek} \nocite{bai2023qwen}\nocite{agarwal2025gpt}

This rapid advancement of LLMs has encouraged social scientists to adopt them as tools for replicating the nuanced interpretive judgments once performed by human coders. Ornstein, Blasingame, and Truscott (2025) reconceptualize LLMs as “co-analysts” capable of human-like interpretation and semantic integration, demonstrating that models can capture discursive and conceptual patterns in political texts rather than relying on superficial lexical cues. Di Leo et al. (2025) further illustrate this potential in the Mapping (A)Ideology project, where GPT-based models estimated the ideological positions of European political parties through pairwise comparisons, producing scales that closely align with expert evaluations. These studies collectively underscore that LLMs can serve as interpretive analytical agents, providing a foundation for our approach, which seeks to evaluate whether such models can reproduce the holistic, meaning-based judgments central to HG.
\nocite{DiLeoEtAl2025} \nocite{OrnsteinEtAl2025}

Moreover, qualitative researchers are now exploring how LLMs can be incorporated into interpretive inquiry. Tai et al. (2024) show that LLMs can apply predefined codebooks to interview data with consistency comparable to human coders, serving as collaborative tools that enhance reliability and efficiency. Schroeder et al. (2024) further find that researchers increasingly treat LLMs as partners in interpretive tasks such as theme identification and code generation, while remaining attentive to ethical and epistemic concerns. It suggests that in qualitative research as well, LLMs are beginning to function not merely as automation tools but as participants in the interpretive process.
\nocite{schroeder2024large} \nocite{tai2024examination}

We therefore contend that LLMs possess the potential to move beyond mere prediction or classification and toward an integrated, human-like understanding of textual meaning. Ideational constructs such as populism, popular sovereignty, and anti-elitism involve latent meaning structures that cannot be captured through word frequencies or surface-level syntax alone. LLMs, by virtue of their statistical and semantic learning capacities, are capable of recognizing these complex discursive patterns, and this capacity is amplified when combined with the HG approach. HG provides a framework that synthesizes human evaluators’ interpretive intuition and normative judgment, allowing for the quantitative assessment of an LLM’s contextual coherence and ideational reasoning. This study examines whether LLMs can reproduce meaning-based judgments at the discourse level rather than the word level. While most prior work has relied on codebook-based classification or zero-shot prompts, our approach aims to teach the reasoning process central to HG. This is not simple automation but an attempt to position LLMs as collaborative analytical agents that bridge computational inference and human interpretation. The next section outlines how we implement this framework and evaluate LLM performance within the HG process.

\section{Methods} 
To replicate the HG technique using artificial intelligence, we followed the established methodology developed by Hawkins (2009) for measuring populist discourse. The HG approach asks trained coders to read entire speeches and assign scores based on their overall assessment of populist content, using a zero to two scale where zero indicates little or no populism, a one represents clear populism, but used inconsistently or with a mild tone, and a two denotes strong populism throughout the text (Hawkins, 2009; Hawkins et al., 2019).
\nocite{hawkins2009measuring,hawkins2019ideational}

\subsection{Training Material Preparation}

We began by transcribing the complete training materials available on Team Populism's website (Hawkins et al., 2019), which consisted of seven anchor speeches originally designed to train human coders on different combinations and intensities of populist elements (Hawkins, 2009). These speeches serve as exemplars across the 0-2 scale, helping coders understand how populist discourse manifests in different contexts and intensities. We adapted this material from its original video format to create a structured training protocol suitable for large language models, streamlining the content while preserving the essential pedagogical elements that teach coders to recognize populist discourse patterns.
\nocite{hawkins2009measuring} \nocite{hawkins2019globalpopulism}

However, the original training set exhibited a left-skewed distribution across the populism scale, with anchor speeches representing scores of 0.0 (Tony Blair and George Bush), 0.3 (Barack Obama), 1.0 (Stephen Harper), 1.5 (Sarah Palin), 1.7 (Robert Mugabe), and 2.0 (Evo Morales). To address this limitation and provide more comprehensive coverage of the scale, we selected and incorporated three additional speeches: 0.5 (Tony Abbott), 0.8 (Ted Cruz), and 1.3 (Andrés Manuel López Obrador). These supplementary anchors were chosen to represent different political contexts and rhetorical styles while filling critical gaps in the scale coverage, particularly in the moderate populism range (0.4-0.9) that was underrepresented in the original training set.

\subsection{AI Implementation Process}

Following the principle that holistic grading requires coders to interpret whole texts rather than count specific words or phrases (Hawkins, 2009), we designed prompts that instruct the AI to evaluate entire speeches using the same theoretical framework applied in human coding. The AI receives the complete set of anchor speeches with their corresponding scores and detailed explanations of what populist elements justify each rating. This mirrors the training process used for human coders, where exposure to exemplar texts helps establish consistent scoring standards (Hawkins et al., 2019).
\nocite{hawkins2009measuring,hawkins2019ideational}

The AI coding process involves presenting each target speech alongside the complete training materials and explicit instructions to assess populist discourse based on this structured reasoning framework. The system evaluates how prominently populist elements appear throughout the speech and assigns scores on the same 0-2 continuum used in traditional holistic grading methodology, with every scoring decision accompanied by detailed justification that researchers can examine to verify theoretical appropriateness.

\subsection{Chain-of-Thought Prompting Implementation}

To replicate the structured reasoning process that human coders employ during holistic grading, we implemented Chain-of-Thought (CoT) prompting, which guides large language models to produce step-by-step reasoning before reaching final conclusions (Wei et al., 2022). CoT prompting proves particularly valuable for evaluative tasks requiring nuanced judgment, as it forces models to make their analytical process explicit rather than providing unexplained classifications (Cohn et al., 2024).
\nocite{wei2022chain,cohn2024cot_assessment}

Our CoT prompt architecture directly mirrors the sequential stages of human coder training through five distinct phases: 

\begin{enumerate}
    \item theoretical foundation, where we provide the ideational definition of populism, contrast it with pluralism, and include four detailed examples that demonstrate how to reason through and identify different elements of populist discourse in short paragraphs; 
    \item methodological instruction, in which we describe holistic grading principles and why this approach suits latent constructs like populism, accompanied by an example; 
    \item rubric training, which details the 0-2 scoring scale with anchor points and outlines six populist categories (Manichaean vision, cosmic proportions, populist notion of people, elite as conspiring evil, systemic change, and anything-goes attitude) alongside six corresponding pluralist categories; 
    \item anchor speech training, which present the ten training speeches with complete reasoning showing how expert coders systematically evaluate people-centric language, anti-elite framing, and Manichaean binaries before synthesizing these elements into overall populist intensity scores; 
    \item and implementation instructions, providing detailed guidance on holistic integration and score assignment procedures.
\end{enumerate}

Each training example within our prompt demonstrates both the final score and the complete reasoning chain that justifies that score, effectively teaching the model not merely what to conclude but how human experts think through the evaluation systematically. This explicit modeling ensures that the AI replicates the deliberative process that characterizes expert human coding rather than relying on superficial textual cues (Hawkins, 2009).
\nocite{hawkins2009measuring}
    
This approach offers significant advantages over alternative prompting strategies. Zero-shot prompting risks superficial analysis based on irrelevant textual cues, while instruction-only prompting provides evaluation criteria but leaves the reasoning process implicit, potentially leading to inconsistent application of standards. Standard few-shot prompting offers scoring examples but fails to demonstrate the analytical process that produces those scores. In our case, CoT prompting addresses these limitations by explicitly modeling the deliberative process that characterizes effective holistic grading, ensuring that the AI's evaluation mirrors the theoretical and methodological foundations of human coding (Cohn et al., 2024; Wei et al., 2022).
\nocite{wei2022chain,cohn2024cot_assessment}

The CoT framework proves particularly well-suited for populism measurement. This is because populism represents a latent construct that cannot be identified through simple keyword detection; rather, it must be inferred from patterns in rhetorical framing and moral positioning (Hawkins, 2009). By compelling the model to outline its reasoning and systematically evaluate each key aspect of populism, CoT prompts ensure comprehensive and rubric-faithful judgments. This structured approach provides three key advantages: First, it addresses the "black box" critique sometimes leveled at holistic grading by making every aspect of the AI's evaluative reasoning transparent and accountable, allowing researchers to verify that scores derive from theoretically appropriate considerations. Second, it ensures consistency in rubric application by building evaluation criteria directly into the reasoning structure, making the model less prone to inconsistent standards or off-topic drift. Finally, it forces the model to make explicit inferences about latent rhetorical patterns rather than relying on surface-level textual features, aligning the evaluation process with the ideational approach to populism while maintaining the transparency and reproducibility essential for scientific replication (Chiang et al. 2025). 
\nocite{chiang2025tract,hawkins2009measuring}

\subsection{Large Language Models}

Our analysis employs a diverse set of state-of-the-art large language models to evaluate the effectiveness of our synthetic holistic grading approach (SHG). We selected models representing different architectural approaches, reasoning capabilities, and organizational strategies to ensure comprehensive assessment of our methodology's robustness across various AI systems. For technical information, see Table~\ref{tab:llm_overview}.

Representing the closed models, we utilized OpenAI's GPT-5, released on August 7, 2025, which represents a unified system combining fast response capabilities with deep reasoning functionality (OpenAI, 2025). The model supports variable reasoning effort through a configurable parameter (minimal, low, medium, high), allowing us to test both rapid evaluation and extended analytical processes. We evaluated GPT-5 in both high reasoning effort mode, which engages the model's chain-of-thought capabilities for complex populism assessment, and minimal reasoning effort mode, which provides faster responses without extensive deliberation.
\nocite{OpenAI2025-GPT5}

As a contrast to the proprietary models, we included four different sets of open-weight models. First, we have OpenAI's open-weight GPT-OSS models. GPT-OSS is a family of Mixture-of-Experts (MoE) Reasoning models, which includes the 120B parameter model (5.1B activated per token) and the 20B parameter model (3.6B activated per token) (OpenAI, 2025b). These state-of-the-art open-weight models were trained with similar methods and techniques OpenAI uses on their frontier models (such as GPT-o3), leading to strong real-world performance that outshines similarly sized open models on reasoning tasks (OpenAI, 2025b). 
\nocite{OpenAI2025b-GPT-oss}

We also included two models from DeepSeekAI, DeepSeek R1, a MoE reasoning model, and DeepSeek V3, a standard MoE model. DeepSeek R1, released in January 2025, features 671 billion parameters with 37 billion activated per token (DeepSeekAI, 2025). It specializes in chain-of-thought reasoning and achieved performance comparable to OpenAI's o-series across mathematics, coding, and reasoning benchmarks (DeepSeekAI, 2025). DeepSeek V3, the standard version, shares the same parameter structure but operates without extended reasoning capabilities.
\nocite{deepseekai2025}

Next, we evaluated Alibaba's Qwen3 235B, released in April 2025, which features 235 billion total parameters with 22 billion activated per token (Qwen Team, 2025). Similar to GPT-OSS models, this is a MoE Reasoning model that uniquely supports both thinking and non-thinking modes through template configuration. The hybrid architecture allows seamless switching between rapid response generation and extended reasoning chains, making it particularly valuable for testing how mode selection affects populism coding accuracy. We tested both operational modes to assess whether thinking-enabled evaluation produces more accurate populism scores.
\nocite{qwen3}

Finally, Meta's Llama 4 family provided our mixture-of-experts baseline models without built-in reasoning capabilities: Llama 4 Maverick and its smaller version, Llama 4 Scout. Llama 4 Maverick features 400 billion total parameters with 17 billion activated through 128 experts, supporting a 1-million-token context window (Meta, 2025). Llama 4 Scout contains 109 billion total parameters with 17 billion activated through 16 experts, offering an industry-leading 10-million-token context window. Both models were released in April 2025 with an August 2024 knowledge cutoff and represent the current state-of-the-art in open-weight multimodal models (Meta, 2025). 
\nocite{metaai2025}

\begin{table}[htbp]
\centering
\small
\caption{Overview of Selected Large Language Models}
\label{tab:llm_overview}
\begin{tabular}{@{}p{2.5cm}p{3.5cm}p{4cm}p{2.5cm}p{2.5cm}@{}}
\toprule
\textbf{Model} & \textbf{Architecture} & \textbf{Parameters / Active per Token} & \textbf{Context Window} & \textbf{Knowledge Cutoff} \\ \midrule
GPT-5 & Hybrid (Reasoning and Standard) & Unknown & 272k & September 30, 2024 \\

GPT-oss family & MoE Reasoning & 120B (5.1B/token); 20B (3.6B/token) & 128k & June, 2024 \\

DeepSeek family & MoE Reasoning (R1); MoE Standard (V3) & 671B (37B/token) & 128k & R1: January, 2025; V3: December, 2024 \\

Qwen3 235B family & MoE Hybrid (Reasoning and Standard) & 235B (22B/token) & 128k & Unknown \\

Llama 4 family & MoE Standard & Maverick: 400B (17B/token); Scout: 109B (17B/token) & Maverick: 1M; Scout: 10M & August, 2024 \\ \bottomrule
\end{tabular}

\smallskip
\raggedright \textit{Sources:} DeepSeekAI (2025), Meta (2025), OpenAI (2025a, 2025b), Qwen Team (2025). 
All models were accessed through OpenRouter, a unified API service that provides access to over 300 AI models through a single endpoint (OpenRouter, 2025).
\end{table}

This diverse model selection enables us to assess whether SHG performance is an artifact of vendor governance (closed-source/proprietary vs. open-weight) or architectural choices (dense “standard”, MoE-standard, MoE reasoning, etc.). The closed/open contrast matters because open-weight releases provide inspectable weights and greater reproducibility and adaptation leverage (fine-tuning, local control and processing), whereas closed systems restrict these levers often in favor of increased performance (Maslej et al., 2025; Sapkota, Roumeliotis \& Karkee, 2025). Architecturally, Mixture-of-Experts (MoE) models use conditional computation – routing each token to a small subset of experts – thereby expanding capacity at roughly constant per-token compute, a mechanism shown to scale efficiently and, under comparable budgets, to match or surpass dense counterparts when properly configured (Fedus, Zoph \& Shazeer, 2021). By contrast, reasoning-tuned models emphasize post-training methods (e.g., reinforced learning-style “deliberate reasoning”) that elicit step-by-step problem solving; recent releases and analysis indicate these procedures substantively shift accuracy on complex tasks independent of raw scale (OpenAI, 2025c; Raschka, 2025; Weng, 2025). Crossing governance with architecture lets us estimate main and interaction effects - e.g., whether openness and task-specific adaptation narrow gaps to proprietary systems and ensembles - so that any SHG improvements can be attributed to generalizable design principles rather than to a single vendor or training recipe (Tang et al., 2025). 
\nocite{maslej2025report,sapkota2025taxonomy}
\nocite{fedus2021scaling}
\nocite{OpenAI2025c-o3o4,raschka2025,weng2025,tang2025opensourcellmscollaborationbeats}

\section{Results} 

We evaluate whether LLMs can replicate Holistic Grading of populism on a 0–2 scale for 12 speeches of leaders in three countries (UK, Turkey, Montenegro; four types: campaign, famous, international, and ribbon cutting) using five independent runs per model (a test-retest framework). We treat the human mean (two coders for UK and Montenegro, and three coders for Turkey) as the benchmark. Results are in Figure 1, which illustrates both the human and AI scores for each speech and country with 99\% bootstrapped confidence intervals (n = 1000)\footnotemark. Figure 1 already foreshadow the results: (1) SHG works – the top reasoning models replicate human HG closely; (2) the best models, Qwen3 235B (Reasoning) and GPT-5 (Reasoning [high]) reach human-level agreement with modest negative bias and mild scale compression; and (3) weaker/effective-capacity models perform substantively worse.

\footnotetext{We report the complete tables with the data displayed on Figure 1 in the Supplementary Material C.}

Below, we evaluate AI scores against human grades in a single step. We assess AI–human agreement through linear and rank associations, intraclass and concordance correlations, and pooled reliability. Additional robustness checks, including error and calibration diagnostics as well as model rankings, are reported in the Appendix.

\begin{figure}[H]  
    \centering
    \includegraphics[width=\textwidth]{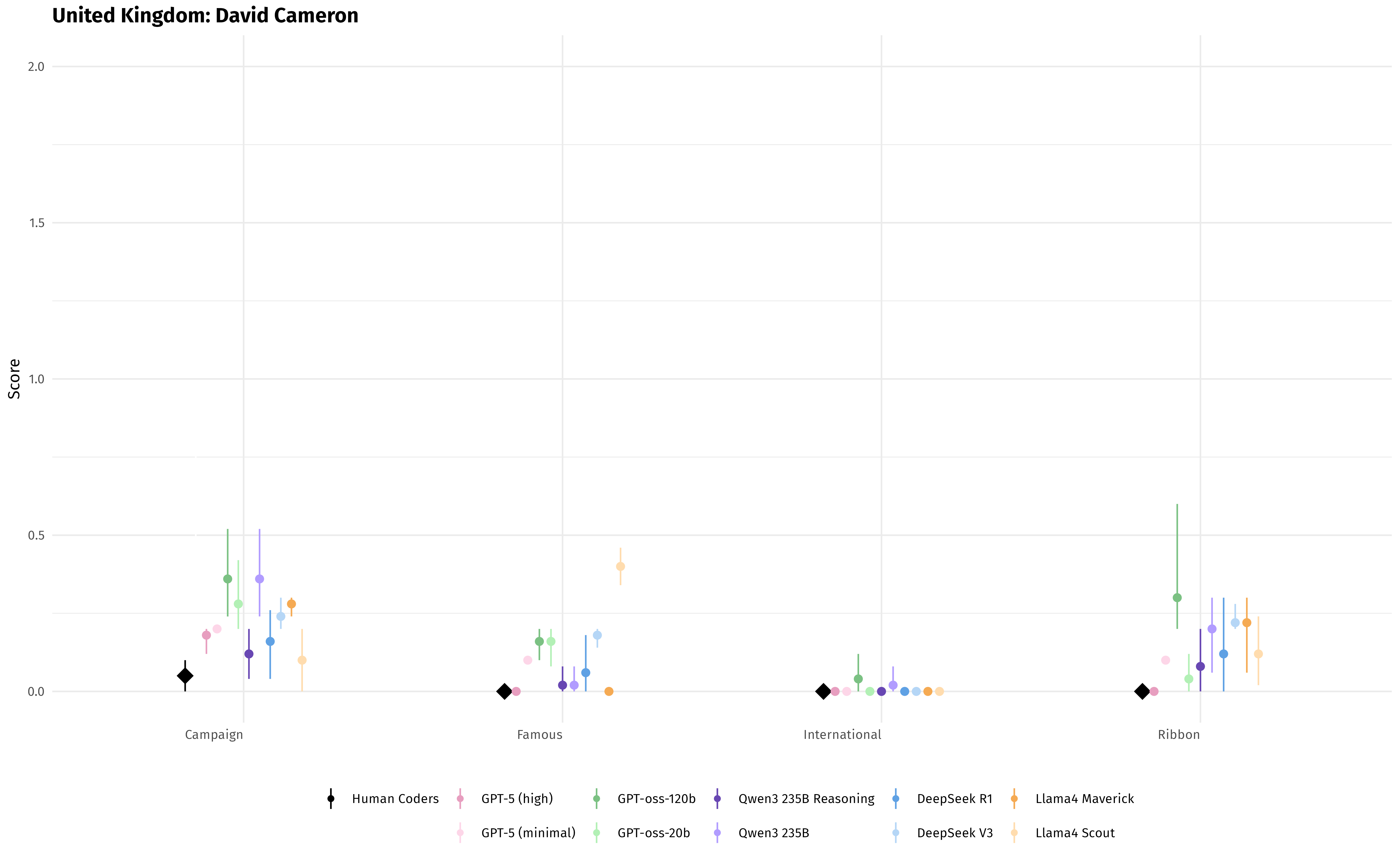}
    \caption{LLM and Human Holistic Grading in the United Kingdom}
    \label{fig:llm_scores}
\end{figure}

\begin{figure}[H]  
    \centering
    \includegraphics[width=\textwidth]{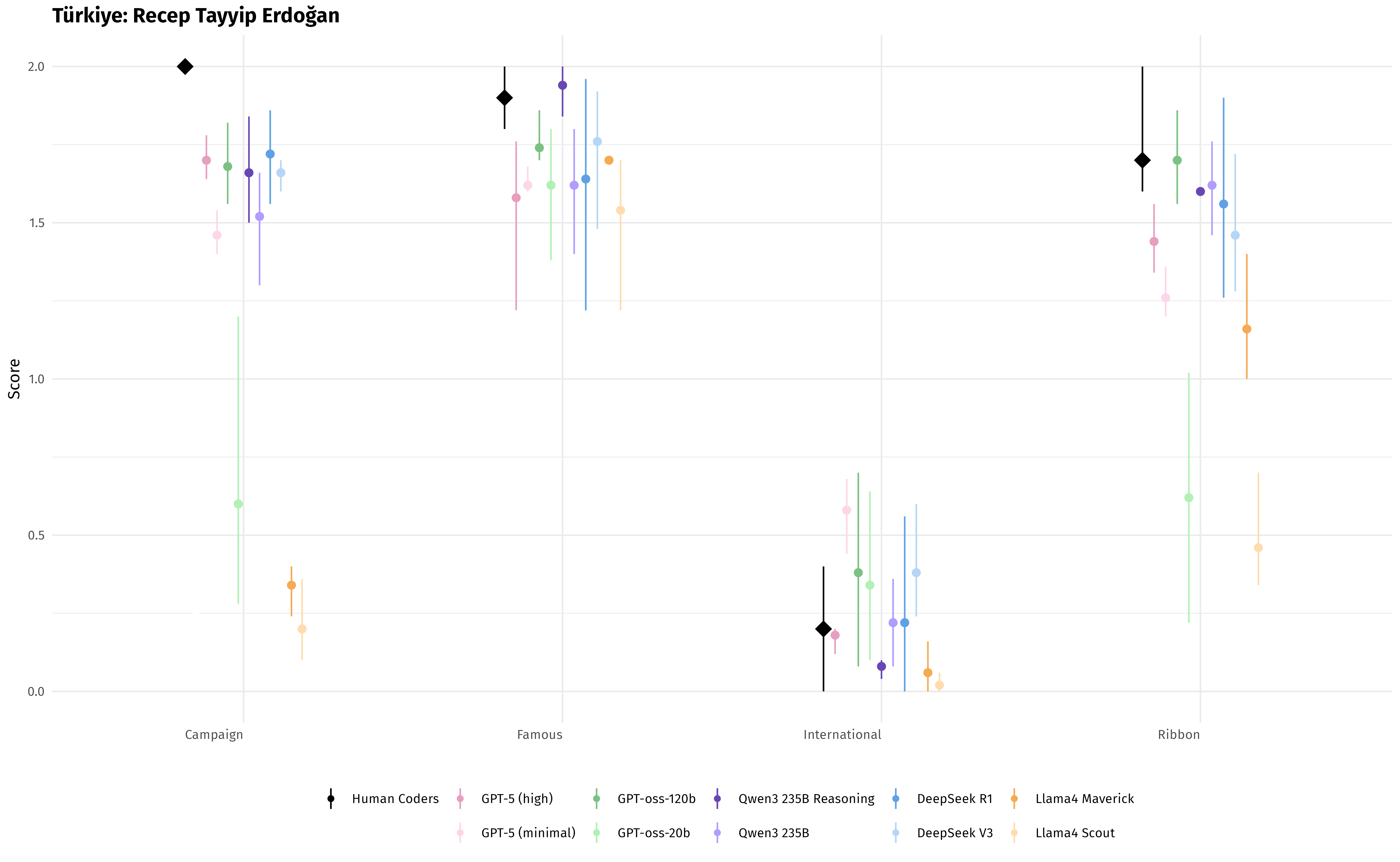}
    \caption{LLM and Human Holistic Grading in Türkiye}
    \label{fig:llm_scores}
\end{figure}

\begin{figure}[H]  
    \centering
    \includegraphics[width=\textwidth]{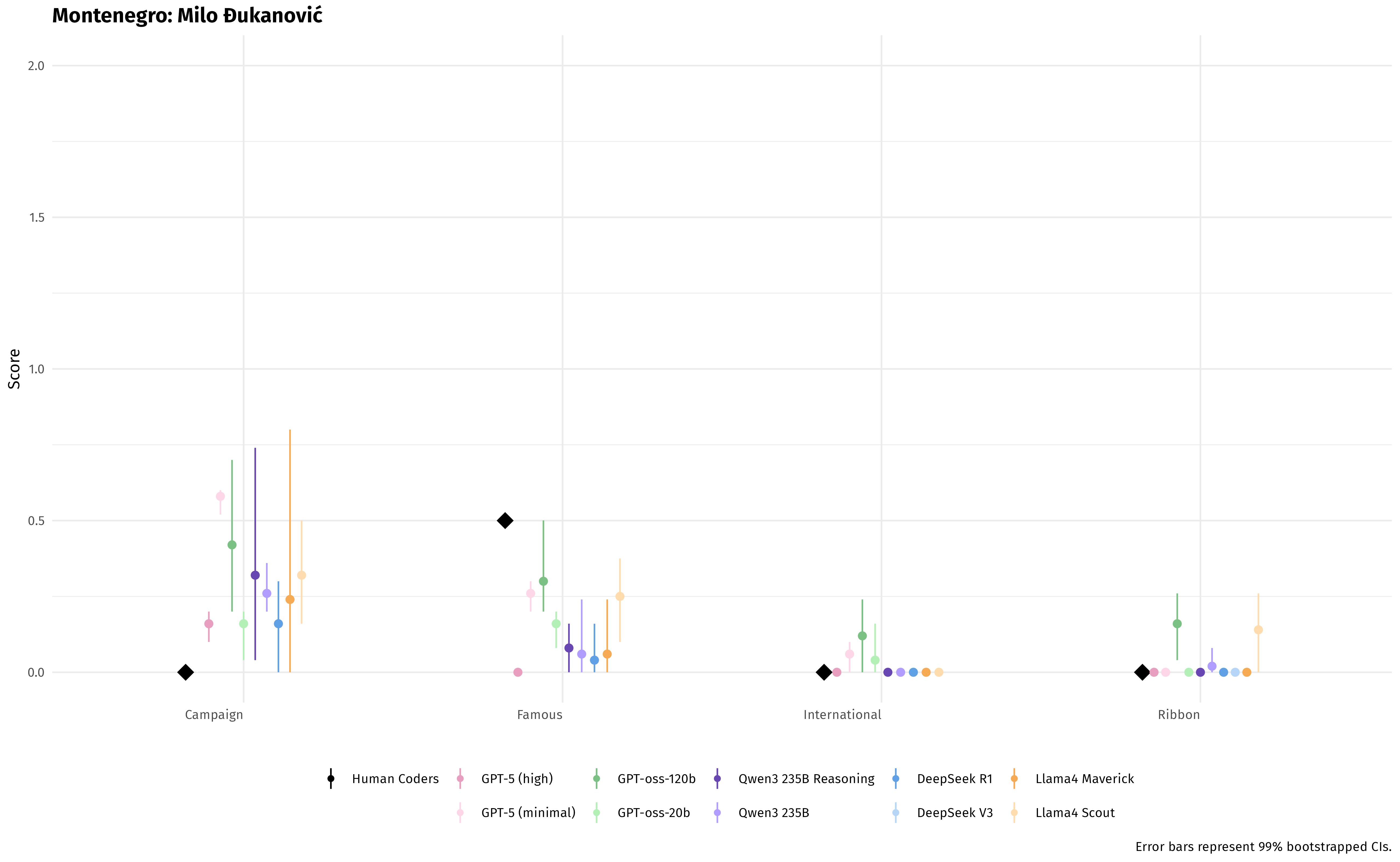}
    \caption{LLM and Human Holistic Grading in Montenegro}

    \label{fig:llm_scores}
\end{figure}

\subsection{AI vs. Human Agreement}

We start by assessing the agreement between AI and human scores model-wise and in four complementary ways. First, we analyze Pearson’s $r$ (linear association) and Spearman’s $\rho$ (rank association), which tell us whether the models track variation across texts, ignoring level offsets or scale shifts. We then calculate intraclass correlation (ICC(2,1), two-way, single, absolute agreement – \textit{aa}), which helps us evaluate interchangeability with the human reference. More simply, ICC answers a strict question: if we treat the human and the AI as two graders, do they give the same numeric score to the same speech? A two-way ICC means we account for differences both across speeches and across raters (human vs. AI), “single” means we judge one rater at a time (a single AI run vs the human), not an average of the many AI runs, and “absolute agreement” (or “\textit{aa}”) means we require exact numerical match – any consistent offset (always \~0.10 higher/lower) or scale mismatch (scores compressed or stretched) count as disagreement. An ICC is equal to 1 only when the AI’s score equals the human’s for every item. Near-zero, or negative ICC, means little to no usable agreement (worse than what one would expect by chance). 

Next, we turn to Lin’s Concordance Correlation (CCC; Lin, 1989). 
Similar to ICC, CCC asks: if we swapped the human score for the AI score, would we basically get the same number for each speech? 
It rewards two things at once: 
(1) The AI must track the human’s ups and downs (high correlation), and 
(2) The AI must be well-calibrated, meaning the same level (average offset, e.g., if the AI is consistently $\sim 0.10$ lower/higher than the human, that is a level bias) and scale (spread, e.g., if the AI compresses or exaggerates highs and lows, its scale is off). 

In a simple fit, we could write it as 
\[
\text{AI} \approx a + b \times \text{Human},
\]
where level bias would show up as $a \neq 0$ and scale mismatch as $b \neq 1$. CCC approaches $1$ only when the best-fit line has $a \approx 0$ and $b \approx 1$. 
If CCC is notably below correlation, the model tracks variation but is miscalibrated 
(offset and/or compression/stretch).

Finally, we turn to Krippendorff's $\alpha$, 
which provides a more stringent test of reliability by assessing whether the agreements we observe are systematic rather than occurring by chance. Unlike the previous metrics that focus on pairwise agreement, Krippendorff's $\alpha$ evaluates the overall consistency of the coding process, treating both human and AI assessments as part of the same measurement system. This metric is particularly valued in content analysis and holistic grading studies (e.g., Bojić et al., 2025; Dzebo et al., 2025; Hawkins and Castanho Silva, 2019) because it accounts for the expected chance agreement, offering a conservative assessment of true consensus.
\nocite{semir2025,bojic2025evaluatinglargelanguagemodels,hawkins2019textual,krippendorff2019content}

Krippendorff's $\alpha$ ranges from $0$ (no agreement beyond chance) 
to $1$ (perfect agreement), with established thresholds: $\alpha \geq 0.8$ indicates reliable data suitable for confident conclusions, while $\alpha$ between $0.667$ and $0.8$ suggests data adequate only for ``drawing tentative conclusions'' (Krippendorff, 2019, 241). We calculate $\alpha$ in two ways: comparing human and AI coders together (treating humans as a single aggregated coder) and examining agreement among AI coders only to assess internal model consistency. Table \ref{tab:model_performance} presents the pairwise agreement metrics (Pearson's $r$, Spearman's $\rho$, ICC(2,1), and CCC), while Table \ref{tab:intercoder_agreement} shows Krippendorff's $\alpha$ calculated both for human--AI agreement and AI-only agreement.
\nocite{krippendorff2019content} 

\begin{table}[htbp]
\footnotesize
\centering
\begin{threeparttable}
\caption{Performance Comparison of Large Language Models}
\label{tab:model_performance}
\begin{tabularx}{\textwidth}{@{}p{1.5cm}p{2.2cm}p{2cm}p{1.5cm}p{1.5cm}p{1.5cm}p{1.5cm}p{1.5cm}@{}}
\toprule
\textbf{Model} & \textbf{Architecture} & \textbf{Pearson’s $r$} & \textbf{Spearman’s $\rho$} & \textbf{ICC} & \textbf{CCC} & \textbf{MAE} & \textbf{RMSE} \\ \midrule
GPT-5 & Reasoning (high) & 
0.974 \newline (0.011) & 0.794 \newline (0.011) & 0.951 \newline (0.026) & 0.947 \newline (0.028) & 0.146 \newline (0.027) & 0.233 \newline (0.050) \\

GPT-5 & Standard (minimal) & 0.943 \newline (0.010) & 0.847 \newline (0.016) & 0.902 \newline (0.012) & 0.894 \newline (0.012) & 0.245 \newline (0.016) & 0.320 \newline (0.017) \\
GPT-oss 120B & MoE Reasoning & 0.952 \newline (0.035) & 0.761 \newline (0.086) & 0.932 \newline (0.045) & 0.926 \newline (0.048) & 0.221 \newline (0.061) & 0.272 \newline (0.079) \\
GPT-oss 20B & MoE Reasoning & 0.761 \newline (0.161) & 0.758 \newline (0.097) & 0.644 \newline (0.169) & 0.626 \newline (0.175) & 0.335 \newline (0.076) & 0.559 \newline (0.159) \\
DeepSeek R1 & MoE Reasoning & 0.953 \newline (0.026) & 0.732 \newline (0.100) & 0.939 \newline (0.046) & 0.934 \newline (0.050) & 0.169 \newline (0.058) & 0.257 \newline (0.086) \\
DeepSeek V3 & MoE Standard & 0.947 \newline (0.020) & 0.786 \newline (0.048) & 0.930 \newline (0.022) & 0.924 \newline (0.024) & 0.209 \newline (0.037) & 0.280 \newline (0.044) \\
Qwen3 235B & MoE Reasoning (high) & 0.961 \newline (0.037) & 0.761 \newline (0.157) & 0.958 \newline (0.031) & 0.954 \newline (0.034) & 0.133 \newline (0.038) & 0.223 \newline (0.079) \\
Qwen3 235B & MoE Standard & 0.953 \newline (0.019) & 0.718 \newline (0.080) & 0.934 \newline (0.024) & 0.929 \newline (0.026) & 0.192 \newline (0.038) & 0.270 \newline (0.049) \\
Llama 4 Maverick & MoE Standard & 0.760 \newline (0.097) & 0.666 \newline (0.166) & 0.690 \newline (0.086) & 0.672 \newline (0.089) & 0.311 \newline (0.062) & 0.556 \newline (0.066) \\
Llama 4 Scout & MoE Standard & 0.608 \newline (0.090) & 0.374 \newline (0.202) & 0.482 \newline (0.052) & 0.460 \newline (0.052) & 0.413 \newline (0.038) & 0.684 \newline (0.030) \\ \bottomrule
\end{tabularx}
\begin{tablenotes}
\footnotesize
\item[a] \textbf{RE: Reasoning Effort} – OpenAI allows users to define the “reasoning effort” when using GPT-5. Reasoning effort guides the model on the number of thinking tokens used when generating a response. While “reasoning effort: high” allocates many tokens for reasoning (values may vary), “reasoning effort: minimal” uses few or none. As of August 2025, OpenAI does not allow users to explicitly turn off reasoning for GPT-5.
\end{tablenotes}
\end{threeparttable}

\end{table}

\begin{table}[htbp]
\footnotesize
\centering
\begin{threeparttable}
\caption{Intercoder Agreement by Model ($\alpha$ All, $\alpha$ AI Only)}
\label{tab:intercoder_agreement}
\begin{tabular}{@{}llccc@{}}
\toprule
\textbf{Model} & \textbf{Architecture} & $\boldsymbol{\alpha}$ \textbf{ All}\tnote{a} & $\boldsymbol{\alpha}$ \textbf{ AI only} & \textbf{Number of Coders} \\ \midrule
GPT-5 & Reasoning (high) & 0.969 & 0.981 & 6 (1 human + 5 AI) \\
GPT-5 & Standard (minimal) & 0.950 & 0.990 & 6 (1 human + 5 AI) \\
GPT-oss 120B & MoE Reasoning & 0.927 & 0.927 & 6 (1 human + 5 AI) \\
GPT-oss 20B & MoE Reasoning & 0.714 & 0.797 & 6 (1 human + 5 AI) \\
DeepSeek R1 & MoE Reasoning & 0.923 & 0.916 & 6 (1 human + 5 AI) \\
DeepSeek V3 & MoE Standard & 0.945 & 0.958 & 6 (1 human + 5 AI) \\
Qwen3 235B & MoE Reasoning (high) & 0.957 & 0.958 & 6 (1 human + 5 AI) \\
Qwen3 235B & MoE Standard & 0.950 & 0.962 & 6 (1 human + 5 AI) \\
Llama 4 Maverick & MoE Standard & 0.804 & 0.914 & 6 (1 human + 5 AI) \\
Llama 4 Scout & MoE Standard & 0.646 & 0.872 & 6 (1 human + 5 AI) \\ \bottomrule
\end{tabular}
\begin{tablenotes}
\footnotesize
\item[a] \textbf{Krippendorff’s $\alpha$ for the human-only grades:} UK $\alpha = 0$, Turkey $\alpha = 0.955$, Montenegro $\alpha = 0$. This happens because Krippendorff’s $\alpha$ is chance-corrected, so when almost every item gets the same value from everyone (e.g., a sea of 0s with a single blip), the observed disagreement ends up about equal to the expected-by-chance disagreement and $\alpha$ stays low or even 0. That does not signal human conflict; it signals too little variation in the data to detect coordination beyond “we mostly use 0.”
\end{tablenotes}
\end{threeparttable}
\end{table}

At the \textit{model} level, the two top reasoning systems demonstrate exceptional agreement with human graders, though with complementary strengths. GPT-5 (reasoning high) achieves CCC $= .947$ and ICC $= .951$ with remarkable stability (CCC and ICC SDs $\approx .026$--$.028$), the highest Pearson correlation ($r = .974$) and a strong Spearman correlation ($\rho = .794$). Qwen3 (reasoning) attains slightly higher absolute agreement metrics (CCC $= .954$; ICC $= .958$) and comparable Pearson correlation ($r = .961$), though with somewhat greater variability across runs (SDs $\approx .031$--$.037$) and lower Spearman correlation ($\rho = .761$). These patterns indicate that both systems largely ``see'' the same ideational signal that human holistic graders detect---achieving high correlation and strong absolute agreement at the system level, with GPT-5 showing superior consistency and Qwen3 showing marginally higher peak performance.

Turning to pooled reliability using Krippendorff's $\alpha$ (interval) with six raters (human mean + five AI runs), the top systems again demonstrate exceptional performance (see Table~3). GPT-5 (reasoning) achieves $\alpha = .969$ (AI-only $\alpha = .981$), while Qwen3 (reasoning) reaches $\alpha = .957$ (AI-only $\alpha = .958$) --- both well above the $\alpha \geq .8$ threshold for reliable data. Weaker open-source models show substantially lower reliability: GPT-oss 20B manages only $\alpha = .714$ and Llama Scout $\alpha = .646$ when combined with the human reference. Notably, these weaker models show higher AI-only $\alpha$ values (e.g., Llama Scout AI-only $\alpha = .872$), suggesting internal consistency among runs while systematic displacement from the human scale---they agree with themselves but deviate in level from human graders.

The pattern of high Pearson/ICC/CCC coupled with relatively lower Spearman correlations (especially for some models) reflects the challenges of rank-order statistics with small N and extensive ties (i.e., repeated values) at the floor (UK and Montenegro speeches cluster near zero), which depresses rank-based measures without necessarily indicating poor absolute agreement. Krippendorff’s $\alpha$ results reinforce our core finding: top reasoning systems integrate smoothly with the human benchmark as reliable additional raters, while weaker models maintain internal consistency but fail to capture the human assessment scale.

On agreement metrics alone, GPT-5 (reasoning) and Qwen3 (reasoning) function as high-quality holistic graders: reliable across runs and demonstrating strong absolute agreement with human scores. Weaker open-source models show meaningful limitations in human-AI agreement despite internal consistency. 

Having established strong agreement between (top) AI systems and human graders, we turn precision and calibration of these models. Full analyses are reported in the Supplementary Material C1 and C2, where we track three error statistics and two calibration diagnostics to assess both the magnitude of AI-human differences and whether those differences follow systematic patters that could undermine practical application.

Our error analysis reveals a clear performance hierarchy among AI models, with top-tier reasoning systems demonstrating substantially superior precision compared to weaker alternatives. The strongest models exhibit small typical errors and near-optimal calibration properties, with disagreements rarely exceeding acceptable margins for content analysis applications. Critically, these systems display systematic rather than random error patterns, including a consistent tendency toward conservative scoring and mild scale compression. While such models occasionally under-score highly populist content and over-score content at the scale's lower end, these biases follow predictable patterns that may reflect a coherent "cautious" scoring philosophy rather than measurement noise.

In stark contrast, weaker open-source models demonstrate qualitatively different performance profiles that render them unsuitable for reliable content analysis. These systems exhibit error magnitudes several times larger than their high-performing counterparts, accompanied by severe miscalibration and inconsistent error patterns across the populism scale. The combination of large absolute errors, systematic under-scoring bias, and unstable calibration properties indicates that weaker models fundamentally fail to approximate human judgment within acceptable research standards. For researchers considering automated holistic grading, these findings suggest that model selection represents a critical methodological decision, with only the most advanced reasoning systems meeting the precision requirements necessary for valid populism measurement.

Because our objective is to test whether AI systems can stand in for human graders in SHG, we close by ranking models using a synthesis of the evidence presented so far rather than a single statistic. Full analyses are once again reported in Supplementary Material C3. In first place, the best overall model is Qwen3 235B (reasoning). Right behind it, we have GPT-5 (reasoning [high]). At the other end of the ranking, next-to-last and at the end of the ranking we have, in order, GPT-oss 20B and Llama 4 Scout. 

Taken together, our results show that automated SHG is feasible with the top reasoning models. Qwen3 235B (open-weight) and GPT-5 achieve near-interchangeability with human graders, with modest negative bias, and mild scale compression. The fact that an open-weight model matches a flagship proprietary system is another intriguing finding, as it matters for auditability, cost, and reproducibility, while it also indicates that researchers can attain human-comparable scoring without relying on closed APIs. By contrast, models with lower effective capacity – GPT-oss 20B and Llama 4 Scout (MoE, 109B parameters with ~17B active) – exhibit larger errors, weaker pooled reliability, poorer calibration, underscoring that effective capacity and the reasoning setup are critical for speech-level fidelity.

\section{Discussion and Conclusion} 

Our study begins from a practical and conceptual problem in the literature. Measuring the ideational content of populism requires sensitivity to context and to the people against elites frame that travels across parties, languages, and time. Human holistic grading remains the benchmark because it reads texts in full and renders an integrated judgment. Yet it is expensive, slow, and difficult to scale. We asked whether a large language model that is prompted with the same materials and rules used for human coders can approximate expert grading while preserving the conceptual richness of the task. We implemented an explicit training and evaluation pipeline that adapts Global Populism Database materials, instructs the model to reason step-by-step (chain of thought), and tests performance on twelve speeches from the United Kingdom, Turkey, and Montenegro with a test-retest design and multiple agreement and calibration metrics.

The evidence shows that structured prompting aligned to holistic grading works. The strongest reasoning models match human coders on both association and absolute agreement, with high Pearson and Spearman correlations, high intraclass correlation, and high concordance. Pooled reliability with Krippendorff’s alpha confirms that the best systems function as reliable additional raters when combined with humans. At the same time, systematic patterns appear. Scores show mild scale compression that lifts values near zero and pulls down values near two, while preserving rank order across speech types. Variation across models matters. GPT 5 with high reasoning effort and Qwen3 235B reasoning deliver the most stable alignment with human graders, while smaller or non-reasoning variants and several open weight baselines are internally consistent but displaced from the human scale.

These results have direct implications for research on populism and for computational text analysis. First, they demonstrate that ideational constructs that are latent and context dependent can be measured with automation when the method reproduces the training, anchors, and integrative logic of holistic grading rather than relying on surface cues. Second, the chain of thought design produces transparent reasoning traces that can be audited and used as teaching material for human coders, which strengthens validity and reproducibility. Third, the approach scales across languages and corpora at low marginal cost, which opens the door to broader comparative coverage and to high-frequency monitoring of executive rhetoric. Finally, the same design can be adapted to adjacent constructs, including pluralist counter framing, nationalist boundary drawing, and crisis or moral threat narratives, while retaining a clear link to established conceptual work.

These strengths notwithstanding, several caveats remain. First, long prompts (~65k tokens) expose the method to potential “context-rot” (Hong, Troynikov and Huber, 2025), wherein performance can degrade as input length grows; frontier models are improving at long-context retention (cf. Fiction.liveBench, 2025), but smaller or older systems remain vulnerable. This is a plausible failure mode for any prompt-heavy setup like ours. 
\nocite{hong2025,fictionlive2025}

Second, chain-of-thought (CoT) is not universally beneficial. Its effectiveness depends on model class and task, and reasoning-tuned models often see only marginal gains relative to the extra time/tokens (Meincke et al., 2025); accordingly, a natural robustness check is to compare against alternative prompting strategies (e.g., direct, few-shot, self-consistency).
\nocite{meincke2025}

Third, large models can exhibit strong “reasoning priors” that dominate predictions regardless of prompt evidence in subjective domains (Chochlakis et al., 2024), implying that CoT may sometimes retrieve pre-existing patterns rather than enable genuinely contextual reasoning. These limitations imply that prompt engineering alone is insufficient for domain-specific, subjective grading; accordingly, our roadmap pairs structured prompting with retrieval-augmented generation (RAG) to ground the model in verifiable evidence and reduce prior-driven drift.
\nocite{chochlakis2024largerlanguagemodelsdont}

Building on these results, our next step is to broaden testing across additional countries and languages, extending coverage beyond the United Kingdom, Turkey, and Montenegro to a wider comparative landscape. We will scale the corpus, include a richer variety of speech genres, and release prompts, code, and benchmark sets so that others can evaluate and refine the approach. By making the workflow accessible and replicable, we hope to encourage a stream of follow on studies that adapt this populism AI framework to new contexts, develop domain-specific calibrations, and explore related constructs such as pluralist framing, nationalist boundary work, and crisis rhetoric. Our aim is to help establish a shared empirical foundation that supports cumulative research on populist discourse on a global scale.

\bibliographystyle{unsrtnat}
\bibliography{references}  

\begin{thebibliography}{68}
\providecommand{\natexlab}[1]{#1}
\providecommand{\url}[1]{\texttt{#1}}
\expandafter\ifx\csname urlstyle\endcsname\relax
  \providecommand{\doi}[1]{doi: #1}\else
  \providecommand{\doi}{doi: \begingroup \urlstyle{rm}\Url}\fi

\bibitem[Hawkins et~al.(2019{\natexlab{a}})Hawkins, Carlin, Littvay, and Rovira~Kaltwasser]{hawkins2019ideational}
Kirk~A. Hawkins, Ryan Carlin, Levente Littvay, and Cristóbal Rovira~Kaltwasser, editors.
\newblock \emph{The Ideational Approach to Populism: Concept, Theory, and Analysis}.
\newblock Extremism and Democracy. Routledge, 2019{\natexlab{a}}.

\bibitem[Mudde(2017)]{mudde2017ideational}
Cas Mudde.
\newblock Populism: An ideational approach.
\newblock In Cristóbal Rovira~Kaltwasser, Paul Taggart, Paulina Ochoa~Espejo, and Pierre Ostiguy, editors, \emph{The Oxford Handbook of Populism}. Oxford University Press, 2017.

\bibitem[Stavrakakis(2024)]{stavrakakis2024populist}
Yannis Stavrakakis.
\newblock \emph{Populist Discourse: Recasting Populism Research}.
\newblock Routledge, 2024.

\bibitem[Weyland(2024)]{weyland2024democracy}
Kurt Weyland.
\newblock \emph{Democracy’s Resilience to Populism’s Threat: Countering Global Alarmism}.
\newblock Cambridge University Press, 2024.
\newblock URL \url{https://doi.org/10.1017/9781009432504}.

\bibitem[Aslanidis(2016)]{aslanidis2016ideology}
Paris Aslanidis.
\newblock Is populism an ideology? a refutation and a new perspective.
\newblock \emph{Political Studies}, 64\penalty0 (1):\penalty0 88--104, 2016.

\bibitem[Armony and Armony(2005)]{armony2005indictments}
Ariel~C. Armony and Victor Armony.
\newblock Indictments, myths, and citizen mobilization in argentina: A discourse analysis.
\newblock \emph{Latin American Politics and Society}, 47\penalty0 (4):\penalty0 27--54, 2005.

\bibitem[Dai and Kustov(2022)]{dai2022campaign}
Yaoyao Dai and Alexander Kustov.
\newblock When do politicians use populist rhetoric? populism as a campaign gamble.
\newblock \emph{Political Communication}, 39\penalty0 (3):\penalty0 383--404, 2022.

\bibitem[Hawkins and Castanho~Silva(2019)]{hawkins2019textual}
Kirk~A. Hawkins and Bruno Castanho~Silva.
\newblock Textual analysis: Big data approaches.
\newblock In Kirk~A. Hawkins, Ryan Carlin, Levente Littvay, and Cristóbal Rovira~Kaltwasser, editors, \emph{The Ideational Approach to Populism: Concept, Theory, and Analysis}, Extremism and Democracy. Routledge, 2019.

\bibitem[Jagers and Walgrave(2007)]{jagers2007communication}
Jan Jagers and Stefaan Walgrave.
\newblock Populism as political communication style: An empirical study of political parties’ discourse in belgium.
\newblock \emph{European Journal of Political Research}, 46\penalty0 (3):\penalty0 319--345, 2007.

\bibitem[Rooduijn and Pauwels(2011)]{rooduijn2011measuring}
Matthijs Rooduijn and Teun Pauwels.
\newblock Measuring populism: Comparing two methods of content analysis.
\newblock \emph{West European Politics}, 34\penalty0 (6):\penalty0 1272--1283, 2011.

\bibitem[Hawkins et~al.(2022)Hawkins, Aguilar, Castanho~Silva, Jenne, Kocijan, and Rovira~Kaltwasser]{hawkins2022global}
Kirk~A. Hawkins, Rosario Aguilar, Bruno Castanho~Silva, Erin~K. Jenne, Bojana Kocijan, and Cristóbal Rovira~Kaltwasser.
\newblock Global populism database.
\newblock Harvard Dataverse, April 2022.
\newblock URL \url{https://doi.org/10.7910/DVN/LFTQEZ}.

\bibitem[Argyle et~al.(2023)Argyle, Busby, Fulda, Gubler, Rytting, and Wingate]{argyle2023simulate}
Lisa~P. Argyle, Ethan~C. Busby, Nancy Fulda, Joshua~R. Gubler, Christopher Rytting, and David Wingate.
\newblock Out of one, many: Using language models to simulate human samples.
\newblock \emph{Political Analysis}, 31\penalty0 (3):\penalty0 337–351, 2023.
\newblock \doi{10.1017/pan.2023.2}.

\bibitem[Tamaki and Littvay(2024)]{tamaki2024chrono}
Eduardo~R. Tamaki and Levente Littvay.
\newblock Chrono-sampling: Generative ai enabled time machine for public opinion data collection.
\newblock PsyArXiv, August 2024.
\newblock URL \url{https://doi.org/10.31234/osf.io/49ags}.

\bibitem[Garvey and Blanchard(2025)]{garvey2025lucid}
Aaron Garvey and Simon~J. Blanchard.
\newblock Generative ai as a research confederate: The lucid methodological framework and toolkit for human-ai interactions research.
\newblock Georgetown McDonough School of Business Research Paper No. 5256150, May 2025.
\newblock URL \url{https://ssrn.com/abstract=5256150}.

\bibitem[Gilardi et~al.(2023)Gilardi, Alizadeh, and Kubli]{gilardi2023GPT}
Fabrizio Gilardi, Meysam Alizadeh, and Ma{\"e}l Kubli.
\newblock Chatgpt outperforms crowd workers for text-annotation tasks.
\newblock \emph{Proceedings of the National Academy of Sciences}, 120\penalty0 (30):\penalty0 e2305016120, 2023.
\newblock \doi{10.1073/pnas.2305016120}.

\bibitem[Ziems et~al.(2024)Ziems, Held, Shaikh, Chen, Zhang, and Yang]{ziems2024transform}
Caleb Ziems, William Held, Omar Shaikh, Jiaao Chen, Zhehao Zhang, and Diyi Yang.
\newblock Can large language models transform computational social science?
\newblock \emph{Computational Linguistics}, 50\penalty0 (1):\penalty0 237--291, March 2024.
\newblock \doi{10.1162/coli_a_00502}.
\newblock URL \url{https://aclanthology.org/2024.cl-1.8/}.

\bibitem[Jenne et~al.(2021)Jenne, Hawkins, and Castanho~Silva]{jenne2021mapping}
Erin~K. Jenne, Kirk~A. Hawkins, and Bruno Castanho~Silva.
\newblock Mapping populism and nationalism in leader rhetoric across north america and europe.
\newblock \emph{Studies in Comparative International Development}, 56:\penalty0 170--196, 2021.

\bibitem[Akkerman et~al.(2017)Akkerman, Zaslove, and Spruyt]{akkerman2017we}
Agnes Akkerman, Andrej Zaslove, and Bram Spruyt.
\newblock {'We the People' or 'We the Peoples'? A Comparison of Support for the Populist Radical Right and Populist Radical Left in the Netherlands}.
\newblock \emph{Swiss Political Science Review}, 23\penalty0 (4):\penalty0 377--403, 2017.

\bibitem[Rooduijn and Akkerman(2017)]{rooduijn2017flank}
Matthijs Rooduijn and Tjitske Akkerman.
\newblock Flank attacks populism and left-right radicalism in western europe.
\newblock \emph{Party Politics}, 23\penalty0 (3):\penalty0 193--204, 2017.

\bibitem[Canovan(1981)]{canovan1981populism}
Margaret Canovan.
\newblock \emph{Populism}.
\newblock Harcourt Brace Jovanovich, 1981.

\bibitem[de~la Torre(2000)]{torre2000seduction}
Carlos de~la Torre.
\newblock \emph{Populist Seduction in Latin America: The Ecuadorian Experience}.
\newblock Ohio University Press, 2000.

\bibitem[Hawkins(2003)]{hawkins2003venezuela}
Kirk~A. Hawkins.
\newblock Populism in venezuela: The rise of chavismo.
\newblock \emph{Third World Quarterly}, 24\penalty0 (6):\penalty0 1137--1160, 2003.

\bibitem[Laclau(1977)]{laclau1977politics}
Ernesto Laclau.
\newblock \emph{Politics and Ideology in Marxist Theory: Capitalism, Fascism, Populism}.
\newblock New Left Books, 1977.

\bibitem[Mudde(2004)]{mudde2004zeitgeist}
Cas Mudde.
\newblock The populist zeitgeist.
\newblock \emph{Government and Opposition}, 39\penalty0 (4):\penalty0 542--563, 2004.

\bibitem[Bonikowski and Gidron(2016)]{bonikowski2016style}
Bart Bonikowski and Noam Gidron.
\newblock The populist style in american politics: Presidential campaign discourse, 1952--1996.
\newblock \emph{Social Forces}, 94\penalty0 (4):\penalty0 1593--1621, 2016.

\bibitem[Lindberg et~al.(2022)Lindberg, Düpont, Higashijima, et~al.]{lindberg2022vparty}
Staffan~I. Lindberg, Nils Düpont, Masaaki Higashijima, et~al.
\newblock Codebook varieties of party identity and organization (v-party) v2. varieties of democracy (v-dem) project.
\newblock \url{https://doi.org/10.23696/VPARTYDSV1}, 2022.

\bibitem[Norris(2020)]{norris2020measuring}
Pippa Norris.
\newblock Measuring populism worldwide.
\newblock \emph{Party Politics}, 26\penalty0 (6):\penalty0 697--717, 2020.

\bibitem[Wiesehomeier et~al.(2021)Wiesehomeier, Singer, and Ruth-Lovell]{wiesehomeier2021prepps}
Nina Wiesehomeier, Matthew Singer, and Saskia Ruth-Lovell.
\newblock Political representation, executives, and political parties survey: Data from expert surveys in 18 latin american countries, 2018-2019. prepps latam v2.
\newblock Harvard Dataverse, February 2021.
\newblock URL \url{https://doi.org/10.7910/DVN/JLOYIJ}.

\bibitem[Zaslove et~al.(2025)Zaslove, Huber, and Meijers]{zaslove2025state}
Andrej Zaslove, Robert~A. Huber, and Maurits~J. Meijers.
\newblock The state of populism: Introducing the 2023 wave of the populism and political parties expert survey.
\newblock \emph{Party Politics}, 2025.
\newblock \doi{10.1177/13540688251361813}.
\newblock URL \url{https://doi.org/10.1177/13540688251361813}.

\bibitem[Meijers and Wiesehomeier(2023)]{meijers2023expert}
Maurits~J. Meijers and Nina Wiesehomeier.
\newblock Expert surveys in party research.
\newblock In Neil Carter, Daniel Keith, Gyda~M. Sindre, and Sofia Vasilopoulou, editors, \emph{The Routledge Handbook of Political Parties}. Routledge, 2023.

\bibitem[Meijers and Zaslove(2021)]{meijers2021measuring}
Maurits~J. Meijers and Andrej Zaslove.
\newblock Measuring populism in political parties: Appraisal of a new approach.
\newblock \emph{Comparative Political Studies}, 54\penalty0 (2):\penalty0 372--407, 2021.

\bibitem[Vaswani et~al.(2017)Vaswani, Shazeer, Parmar, Uszkoreit, Jones, Gomez, Kaiser, and Polosukhin]{VaswaniEtAl2017}
Ashish Vaswani, Noam Shazeer, Niki Parmar, Jakob Uszkoreit, Llion Jones, Aidan~N Gomez, {\L}ukasz Kaiser, and Illia Polosukhin.
\newblock Attention is all you need.
\newblock \emph{Advances in neural information processing systems}, 30, 2017.

\bibitem[Naveed et~al.(2025)Naveed, Khan, Qiu, Saqib, Anwar, Usman, Akhtar, Barnes, and Mian]{naveed2025comprehensive}
Humza Naveed, Asad~Ullah Khan, Shi Qiu, Muhammad Saqib, Saeed Anwar, Muhammad Usman, Naveed Akhtar, Nick Barnes, and Ajmal Mian.
\newblock A comprehensive overview of large language models.
\newblock \emph{ACM Transactions on Intelligent Systems and Technology}, 16\penalty0 (5):\penalty0 1--72, 2025.

\bibitem[Mu et~al.(2024)Mu, Dong, Bontcheva, and Song]{mu2024large}
Yida Mu, Chun Dong, Kalina Bontcheva, and Xingyi Song.
\newblock Large language models offer an alternative to the traditional approach of topic modelling.
\newblock \emph{arXiv preprint arXiv:2403.16248}, 2024.

\bibitem[Wei et~al.(2022)]{wei2022chain}
Jason Wei et~al.
\newblock Chain-of-thought prompting elicits reasoning in large language models.
\newblock In \emph{Advances in Neural Information Processing Systems (NeurIPS)}, 2022.
\newblock URL \url{https://arxiv.org/abs/2201.11903}.

\bibitem[Riquelme et~al.(2021)Riquelme, Puigcerver, Mustafa, Neumann, Jenatton, Susano~Pinto, Keysers, and Houlsby]{RiquelmeEtAl2021}
Carlos Riquelme, Joan Puigcerver, Basil Mustafa, Maxim Neumann, Rodolphe Jenatton, Andr{\'e} Susano~Pinto, Daniel Keysers, and Neil Houlsby.
\newblock Scaling vision with sparse mixture of experts.
\newblock \emph{Advances in Neural Information Processing Systems}, 34:\penalty0 8583--8595, 2021.

\bibitem[Zhao et~al.(2024)Zhao, Zhang, Chen, Kawaguchi, and Bing]{zhaoetal2024large}
Yiran Zhao, Wenxuan Zhang, Guizhen Chen, Kenji Kawaguchi, and Lidong Bing.
\newblock How do large language models handle multilingualism?
\newblock \emph{Advances in Neural Information Processing Systems}, 37:\penalty0 15296--15319, 2024.

\bibitem[Manchanda et~al.(2024)Manchanda, Boettcher, Westphalen, and Jasser]{ManchandaEtAl2024}
Jiya Manchanda, Laura Boettcher, Matheus Westphalen, and Jasser Jasser.
\newblock The open source advantage in large language models (llms).
\newblock \emph{arXiv preprint arXiv:2412.12004}, 2024.

\bibitem[Guo et~al.(2025)Guo, Yang, Zhang, Song, Zhang, Xu, Zhu, Ma, Wang, Bi, et~al.]{guo2025deepseek}
Daya Guo, Dejian Yang, Haowei Zhang, Junxiao Song, Ruoyu Zhang, Runxin Xu, Qihao Zhu, Shirong Ma, Peiyi Wang, Xiao Bi, et~al.
\newblock Deepseek-r1: Incentivizing reasoning capability in llms via reinforcement learning.
\newblock \emph{arXiv preprint arXiv:2501.12948}, 2025.

\bibitem[Bai et~al.(2023)Bai, Bai, Chu, Cui, Dang, Deng, Fan, Ge, Han, Huang, et~al.]{bai2023qwen}
Jinze Bai, Shuai Bai, Yunfei Chu, Zeyu Cui, Kai Dang, Xiaodong Deng, Yang Fan, Wenbin Ge, Yu~Han, Fei Huang, et~al.
\newblock Qwen technical report.
\newblock \emph{arXiv preprint arXiv:2309.16609}, 2023.

\bibitem[Agarwal et~al.(2025)Agarwal, Ahmad, Ai, Altman, Applebaum, Arbus, Arora, Bai, Baker, Bao, et~al.]{agarwal2025gpt}
Sandhini Agarwal, Lama Ahmad, Jason Ai, Sam Altman, Andy Applebaum, Edwin Arbus, Rahul~K Arora, Yu~Bai, Bowen Baker, Haiming Bao, et~al.
\newblock gpt-oss-120b \& gpt-oss-20b model card.
\newblock \emph{arXiv preprint arXiv:2508.10925}, 2025.

\bibitem[Di~Leo et~al.(2025)Di~Leo, Zeng, Dinas, and Tamtam]{DiLeoEtAl2025}
Riccardo Di~Leo, Chen Zeng, Elias Dinas, and Reda Tamtam.
\newblock Mapping (a) ideology: A taxonomy of european parties using generative llms as zero-shot learners.
\newblock \emph{Political Analysis}, pages 1--8, 2025.

\bibitem[Ornstein et~al.(2025)Ornstein, Blasingame, and Truscott]{OrnsteinEtAl2025}
Joseph~T Ornstein, Elise~N Blasingame, and Jake~S Truscott.
\newblock How to train your stochastic parrot: Large language models for political texts.
\newblock \emph{Political Science Research and Methods}, 13\penalty0 (2):\penalty0 264--281, 2025.

\bibitem[Schroeder et~al.(2024)Schroeder, Aubin Le~Qu{\'e}r{\'e}, Randazzo, Mimno, and Schoenebeck]{schroeder2024large}
Hope Schroeder, Marianne Aubin Le~Qu{\'e}r{\'e}, Casey Randazzo, David Mimno, and Sarita Schoenebeck.
\newblock Large language models in qualitative research: Can we do the data justice?
\newblock \emph{arXiv e-prints}, pages arXiv--2410, 2024.

\bibitem[Tai et~al.(2024)Tai, Bentley, Xia, Sitt, Fankhauser, Chicas-Mosier, and Monteith]{tai2024examination}
Robert~H Tai, Lillian~R Bentley, Xin Xia, Jason~M Sitt, Sarah~C Fankhauser, Ana~M Chicas-Mosier, and Barnas~G Monteith.
\newblock An examination of the use of large language models to aid analysis of textual data.
\newblock \emph{International Journal of Qualitative Methods}, 23:\penalty0 16094069241231168, 2024.

\bibitem[Hawkins(2009)]{hawkins2009measuring}
Kirk~A. Hawkins.
\newblock Is chávez populist?: Measuring populist discourse in comparative perspective.
\newblock \emph{Comparative Political Studies}, 42\penalty0 (8):\penalty0 1040--1067, 2009.

\bibitem[Hawkins et~al.(2019{\natexlab{b}})Hawkins, Aguilar, Castanho~Silva, Jenne, Kocijan, and Rovira~Kaltwasser]{hawkins2019globalpopulism}
Kirk~A. Hawkins, Rosario Aguilar, Bruno Castanho~Silva, Erin~K. Jenne, Bojana Kocijan, and Cristobal Rovira~Kaltwasser.
\newblock Global populism database: Methodology.
\newblock Team Populism Report, 2019{\natexlab{b}}.
\newblock Available at: \url{https://populism.byu.edu/App_Data/Publications/Global%20Populism%20Database%20Methodology.pdf}.

\bibitem[Cohn et~al.(2024)]{cohn2024cot_assessment}
Clayton Cohn et~al.
\newblock A chain-of-thought prompting approach with llms for evaluating students’ formative assessment responses in science.
\newblock \emph{arXiv preprint}, 2024.
\newblock URL \url{https://arxiv.org/abs/2403.14565}.

\bibitem[Chiang et~al.(2025)]{chiang2025tract}
Cheng-Han Chiang et~al.
\newblock Tract: Regression-aware fine-tuning meets chain-of-thought reasoning for llm-as-a-judge.
\newblock \emph{arXiv preprint}, 2025.
\newblock URL \url{https://arxiv.org/abs/2503.04381}.

\bibitem[{OpenAI}(2025{\natexlab{a}})]{OpenAI2025-GPT5}
{OpenAI}.
\newblock Introducing gpt-5, 2025{\natexlab{a}}.
\newblock URL \url{https://openai.com/index/introducing-gpt-5/}.
\newblock Product release blog.

\bibitem[{OpenAI}(2025{\natexlab{b}})]{OpenAI2025b-GPT-oss}
{OpenAI}.
\newblock Introducing gpt-oss, 2025{\natexlab{b}}.
\newblock URL \url{https://openai.com/index/introducing-gpt-oss/}.
\newblock Product release blog.

\bibitem[DeepSeek-AI et~al.(2025)DeepSeek-AI, Liu, Feng, Xue, Wang, Wu, Lu, Zhao, Deng, Zhang, Ruan, Dai, Guo, Yang, Chen, Ji, Li, Lin, Dai, Luo, Hao, Chen, Li, Zhang, Bao, Xu, Wang, Zhang, Ding, Xin, Gao, Li, Qu, Cai, Liang, Guo, Ni, Li, Wang, Chen, Chen, Yuan, Qiu, Li, Song, Dong, Hu, Gao, Guan, Huang, Yu, Wang, Zhang, Xu, Xia, Zhao, Wang, Zhang, Li, Wang, Zhang, Zhang, Tang, Li, Tian, Huang, Wang, Zhang, Wang, Zhu, Chen, Du, Chen, Jin, Ge, Zhang, Pan, Wang, Xu, Zhang, Chen, Li, Lu, Zhou, Chen, Wu, Ye, Ye, Ma, Wang, Zhou, Yu, Zhou, Pan, Wang, Yun, Pei, Sun, Xiao, Zeng, Zhao, An, Liu, Liang, Gao, Yu, Zhang, Li, Jin, Wang, Bi, Liu, Wang, Shen, Chen, Zhang, Chen, Nie, Sun, Wang, Cheng, Liu, Xie, Liu, Yu, Song, Shan, Zhou, Yang, Li, Su, Lin, Li, Wang, Wei, Zhu, Zhang, Xu, Xu, Huang, Li, Zhao, Sun, Li, Wang, Yu, Zheng, Zhang, Shi, Xiong, He, Tang, Piao, Wang, Tan, Ma, Liu, Guo, Wu, Ou, Zhu, Wang, Gong, Zou, He, Zha, Xiong, Ma, Yan, Luo, You, Liu, Zhou, Wu, Ren, Ren, Sha, Fu, Xu, Huang, Zhang, Xie, Zhang, Hao,
  Gou, Ma, Yan, Shao, Xu, Wu, Zhang, Li, Gu, Zhu, Liu, Li, Xie, Song, Gao, and Pan]{deepseekai2025}
DeepSeek-AI, Aixin Liu, Bei Feng, Bing Xue, Bingxuan Wang, Bochao Wu, Chengda Lu, Chenggang Zhao, Chengqi Deng, Chenyu Zhang, Chong Ruan, Damai Dai, Daya Guo, Dejian Yang, Deli Chen, Dongjie Ji, Erhang Li, Fangyun Lin, Fucong Dai, Fuli Luo, Guangbo Hao, Guanting Chen, Guowei Li, H.~Zhang, Han Bao, Hanwei Xu, Haocheng Wang, Haowei Zhang, Honghui Ding, Huajian Xin, Huazuo Gao, Hui Li, Hui Qu, J.~L. Cai, Jian Liang, Jianzhong Guo, Jiaqi Ni, Jiashi Li, Jiawei Wang, Jin Chen, Jingchang Chen, Jingyang Yuan, Junjie Qiu, Junlong Li, Junxiao Song, Kai Dong, Kai Hu, Kaige Gao, Kang Guan, Kexin Huang, Kuai Yu, Lean Wang, Lecong Zhang, Lei Xu, Leyi Xia, Liang Zhao, Litong Wang, Liyue Zhang, Meng Li, Miaojun Wang, Mingchuan Zhang, Minghua Zhang, Minghui Tang, Mingming Li, Ning Tian, Panpan Huang, Peiyi Wang, Peng Zhang, Qiancheng Wang, Qihao Zhu, Qinyu Chen, Qiushi Du, R.~J. Chen, R.~L. Jin, Ruiqi Ge, Ruisong Zhang, Ruizhe Pan, Runji Wang, Runxin Xu, Ruoyu Zhang, Ruyi Chen, S.~S. Li, Shanghao Lu, Shangyan Zhou, Shanhuang
  Chen, Shaoqing Wu, Shengfeng Ye, Shengfeng Ye, Shirong Ma, Shiyu Wang, Shuang Zhou, Shuiping Yu, Shunfeng Zhou, Shuting Pan, T.~Wang, Tao Yun, Tian Pei, Tianyu Sun, W.~L. Xiao, Wangding Zeng, Wanjia Zhao, Wei An, Wen Liu, Wenfeng Liang, Wenjun Gao, Wenqin Yu, Wentao Zhang, X.~Q. Li, Xiangyue Jin, Xianzu Wang, Xiao Bi, Xiaodong Liu, Xiaohan Wang, Xiaojin Shen, Xiaokang Chen, Xiaokang Zhang, Xiaosha Chen, Xiaotao Nie, Xiaowen Sun, Xiaoxiang Wang, Xin Cheng, Xin Liu, Xin Xie, Xingchao Liu, Xingkai Yu, Xinnan Song, Xinxia Shan, Xinyi Zhou, Xinyu Yang, Xinyuan Li, Xuecheng Su, Xuheng Lin, Y.~K. Li, Y.~Q. Wang, Y.~X. Wei, Y.~X. Zhu, Yang Zhang, Yanhong Xu, Yanhong Xu, Yanping Huang, Yao Li, Yao Zhao, Yaofeng Sun, Yaohui Li, Yaohui Wang, Yi~Yu, Yi~Zheng, Yichao Zhang, Yifan Shi, Yiliang Xiong, Ying He, Ying Tang, Yishi Piao, Yisong Wang, Yixuan Tan, Yiyang Ma, Yiyuan Liu, Yongqiang Guo, Yu~Wu, Yuan Ou, Yuchen Zhu, Yuduan Wang, Yue Gong, Yuheng Zou, Yujia He, Yukun Zha, Yunfan Xiong, Yunxian Ma, Yuting Yan, Yuxiang
  Luo, Yuxiang You, Yuxuan Liu, Yuyang Zhou, Z.~F. Wu, Z.~Z. Ren, Zehui Ren, Zhangli Sha, Zhe Fu, Zhean Xu, Zhen Huang, Zhen Zhang, Zhenda Xie, Zhengyan Zhang, Zhewen Hao, Zhibin Gou, Zhicheng Ma, Zhigang Yan, Zhihong Shao, Zhipeng Xu, Zhiyu Wu, Zhongyu Zhang, Zhuoshu Li, Zihui Gu, Zijia Zhu, Zijun Liu, Zilin Li, Ziwei Xie, Ziyang Song, Ziyi Gao, and Zizheng Pan.
\newblock Deepseek-v3 technical report, 2025.
\newblock URL \url{https://arxiv.org/abs/2412.19437}.

\bibitem[Team(2025)]{qwen3}
Qwen Team.
\newblock Qwen3 technical report, 2025.
\newblock URL \url{https://arxiv.org/abs/2505.09388}.

\bibitem[AI(2025)]{metaai2025}
Meta AI.
\newblock The llama 4 herd: The beginning of a new era of natively multimodal ai innovation, 2025.
\newblock URL \url{https://ai.meta.com/blog/llama-4-multimodal-intelligence/}.

\bibitem[Maslej et~al.(2025)Maslej, Fattorini, Perrault, Gil, Parli, Kariuki, Capstick, Reuel, Brynjolfsson, Etchemendy, Ligett, Lyons, Manyika, Niebles, Shoham, Wald, Walsh, Hamrah, Santarlasci, and Oak]{maslej2025report}
Nestor Maslej, Loredana Fattorini, Raymond Perrault, Yolanda Gil, Vanessa Parli, Njenga Kariuki, Emily Capstick, Anka Reuel, Erik Brynjolfsson, John Etchemendy, Katrina Ligett, Terah Lyons, James Manyika, Juan~Carlos Niebles, Yoav Shoham, Russell Wald, Tobi Walsh, Armin Hamrah, Lapo Santarlasci, and Sukrut Oak.
\newblock Artificial intelligence index report 2025, 04 2025.

\bibitem[Sapkota et~al.(2025)Sapkota, Roumeliotis, and Karkee]{sapkota2025taxonomy}
Ranjan Sapkota, Konstantinos Roumeliotis, and Manoj Karkee.
\newblock Ai agents vs. agentic ai: A conceptual taxonomy, applications and challenges, 05 2025.

\bibitem[Fedus et~al.(2021)Fedus, Zoph, and Shazeer]{fedus2021scaling}
William Fedus, Barret Zoph, and Noam Shazeer.
\newblock Switch transformers: Scaling to trillion parameter models with simple and efficient sparsity, 01 2021.

\bibitem[{OpenAI}(2025{\natexlab{c}})]{OpenAI2025c-o3o4}
{OpenAI}.
\newblock Introducing openai o3 and o4-mini, 2025{\natexlab{c}}.
\newblock URL \url{https://openai.com/index/introducing-o3-and-o4-mini/?utm_source=chatgpt.com}.
\newblock Product release blog.

\bibitem[Raschka(2025)]{raschka2025}
Sebastian Raschka.
\newblock The state of reinforcement learning for llm reasoning, 2025.
\newblock URL \url{https://magazine.sebastianraschka.com/p/the-state-of-llm-reasoning-model-training?utm_source=chatgpt.com}.

\bibitem[Weng(2025)]{weng2025}
Lilian Weng.
\newblock Why we think, 2025.
\newblock URL \url{https://lilianweng.github.io/posts/2025-05-01-thinking/?utm_source=chatgpt.com}.

\bibitem[Tang et~al.(2025)Tang, Cao, Lin, Hong, Zhang, Hu, Bai, Chen, Ouyang, and Ye]{tang2025opensourcellmscollaborationbeats}
Shengji Tang, Jianjian Cao, Weihao Lin, Jiale Hong, Bo~Zhang, Shuyue Hu, Lei Bai, Tao Chen, Wanli Ouyang, and Peng Ye.
\newblock Open-source llms collaboration beats closed-source llms: A scalable multi-agent system, 2025.
\newblock URL \url{https://arxiv.org/abs/2507.14200}.

\bibitem[Dzebo et~al.(0)Dzebo, Jenne, Littvay, Hawkins, and van~der Veen]{semir2025}
Semir Dzebo, Erin~K. Jenne, Levente Littvay, Kirk~A. Hawkins, and Olaf van~der Veen.
\newblock Us governors populism database: Assessing the impact of donald trump on state-level discourse.
\newblock \emph{Party Politics}, 0\penalty0 (0):\penalty0 13540688251327564, 0.
\newblock \doi{10.1177/13540688251327564}.
\newblock URL \url{https://doi.org/10.1177/13540688251327564}.

\bibitem[Bojic et~al.(2025)Bojic, Zagovora, Zelenkauskaite, Vukovic, Cabarkapa, Jerkovic, and Jovančevic]{bojic2025evaluatinglargelanguagemodels}
Ljubisa Bojic, Olga Zagovora, Asta Zelenkauskaite, Vuk Vukovic, Milan Cabarkapa, Selma~Veseljević Jerkovic, and Ana Jovančevic.
\newblock Evaluating large language models against human annotators in latent content analysis: Sentiment, political leaning, emotional intensity, and sarcasm, 2025.
\newblock URL \url{https://arxiv.org/abs/2501.02532}.

\bibitem[Krippendorff(2019)]{krippendorff2019content}
Klaus Krippendorff.
\newblock \emph{Content Analysis: An Introduction to Its Methodology}.
\newblock SAGE Publications, Inc., Thousand Oaks, CA, 4 edition, 2019.
\newblock \doi{10.4135/9781071878781}.
\newblock URL \url{https://doi.org/10.4135/9781071878781}.

\bibitem[Hong et~al.(2025)Hong, Troynikov, and Huber]{hong2025}
Kelly Hong, Anton Troynikov, and Jeff Huber.
\newblock Context rot: How increasing input tokens impacts llm performance, 2025.
\newblock URL \url{https://research.trychroma.com/context-rot}.

\bibitem[Fiction.live(2025)]{fictionlive2025}
Fiction.live.
\newblock Fiction.livebench august 21 2025, 2025.
\newblock URL \url{https://fiction.live/stories/Fiction-liveBench-Feb-21-2025/oQdzQvKHw8JyXbN87}.

\bibitem[Meincke et~al.(2025)Meincke, Mollick, Mollick, and Shapiro]{meincke2025}
Lennart Meincke, Ethan~R. Mollick, Lilach Mollick, and Dan Shapiro.
\newblock Prompting science report 2: The decreasing value of chain of thought in prompting.
\newblock \emph{SSRN}, 2025.
\newblock \doi{http://dx.doi.org/10.2139/ssrn.5285532}.

\bibitem[Chochlakis et~al.(2024)Chochlakis, Pandiyan, Lerman, and Narayanan]{chochlakis2024largerlanguagemodelsdont}
Georgios Chochlakis, Niyantha~Maruthu Pandiyan, Kristina Lerman, and Shrikanth Narayanan.
\newblock Larger language models don't care how you think: Why chain-of-thought prompting fails in subjective tasks, 2024.
\newblock URL \url{https://arxiv.org/abs/2409.06173}.

\end{thebibliography}






\newpage
\appendix

\clearpage
\thispagestyle{empty}
\setcounter{page}{1}
\renewcommand{\thepage}{S\arabic{page}}
\setcounter{figure}{0}\renewcommand{\thefigure}{S\arabic{figure}}
\setcounter{table}{0}\renewcommand{\thetable}{S\arabic{table}}
\setcounter{equation}{0}\renewcommand{\theequation}{S\arabic{equation}}
\setcounter{footnote}{0}

\begin{center}
  {\LARGE Supplementary Material}\\[0.8em]
  {\Large Populism Meets AI: Advancing Populism Research with LLMs}\\[1.2em]
  
  {
  Eduardo Ryô Tamaki\footnote{German Institute for Global and Area Studies}, 
  Yujin J. Jung\footnote{Mount St. Mary's University}, 
  Julia Chatterley\footnote{Princeton University}, 
  Semir Dzebo\footnote{Central European University}, \\
  Levente Littvay\footnote{ELTE Centre for Social Sciences}, 
  Grant Mitchell\footnote{University of California, Los Angeles}, 
  Cristóbal Sandoval\footnote{Diego Portales University}, 
  Kirk A. Hawkins\footnote{Brigham Young University}\\ [2.5em] }
  
  \begin{tabular}{@{}p{0.7\textwidth}@{}}
      \toprule
      \textbf{Contents} \hfill \textbf{Page}\\\midrule
      Training and Test Set \dotfill \pageref{app:train-test} \\
      Full Tables - Populism Score \dotfill \pageref{app:full-tables} \\
      Error and Calibration \dotfill \pageref{app:error-calibration} \\
      \quad Error Metrics \dotfill \pageref{app:error} \\
      \quad Calibration Diagnostics \dotfill \pageref{app:calibration} \\
      Leaderboard \dotfill \pageref{app:leader} \\
      \bottomrule
  \end{tabular}
  \vfill
\end{center}
\clearpage
\section{Training and Test Set}\label{app:train-test}

\begin{table}[h]
\centering
\caption{Training Set}
\begin{tabular}{lllll}
\toprule
No & Country & Name & Speech Type & Score \\
\midrule
1 & UK & Blair & {\footnotesize European Bank for Reconstruction and Development Meeting} & 0.0 \\
2 & USA & Bush & {\small Address to Joint Session of Congress after 9/11} & 0.0 \\
3 & USA & Obama & {\small State of the Union Address} & 0.3 \\
4 & Australia & Abbott & {\small Campaign Launch} & 0.5 \\
5 & USA & Cruz & {\small Campaign Stop} & 0.8 \\
6 & Canada & Harper & {\small Federal Accountability Act Address} & 1.0 \\
7 & Mexico & Obrador & {\small Inauguration} & 1.3 \\
8 & USA & Palin & {\small Tea Party Convention} & 1.5 \\
9 & Zimbabwe & Mugabe & {\small World Summit on Sustainable Development (WSSD)} & 1.7 \\ 
10 & Bolivia & Morales & {\small Post-uprising Speech} & 2.0 \\

\bottomrule
\end{tabular}
\end{table}

\begin{table}[h]
\centering
\caption{Test Set}
\begin{tabular}{llll}
\toprule
Country & Name & Speech Type & GPD Score \\
\midrule
Montenegro & Djukanovic & Campaign & 0 \\
Montenegro & Djukanovic & Famous & 0.5 \\
Montenegro & Djukanovic & International & 0 \\
Montenegro & Djukanovic & Ribbon & 0 \\
Turkey & Erdogan & Campaign & 2  \\
Turkey & Erdogan & Famous & 1.93  \\
Turkey & Erdogan & International & 0.2  \\
Turkey & Erdogan & Ribbon & 1.73  \\
\rowcolor{lightgray} UK & Cameron & Campaign & 0.05  \\
\rowcolor{lightgray} UK & Cameron & Famous & 0  \\
\rowcolor{lightgray} UK & Cameron & International & 0 \\
\rowcolor{lightgray} UK & Cameron & Ribbon & 0  \\
\bottomrule
\end{tabular}
\end{table}

\newpage
\section{Full Tables - Populism Score}\label{app:full-tables}

\begin{table}[!htbp]
\centering
\caption{Populism Score: United Kingdom – David Cameron}
\label{tab:populism_scores_uk}
\begin{adjustbox}{max width=\textwidth}
\begin{tabular}{llrrrrrlrrrrr}
\toprule
\textbf{Model} & \textbf{Type} & \textbf{Run 1} & \textbf{Run 2} & \textbf{Run 3} & \textbf{Run 4} & \textbf{Run 5} & \textbf{Speech} & \textbf{Mean} & \makecell{\textbf{95\% CI}\\\textbf{Low}} & \makecell{\textbf{95\% CI}\\\textbf{High}} & \makecell{\textbf{99\% CI}\\\textbf{Low}} & \makecell{\textbf{99\% CI}\\\textbf{High}} \\
\midrule

 Deepseek       & noreasoning &    0.30 &    0.20 &    0.20 &    0.20 &    0.30 & campaign      &   0.24 &           0.20 &            0.28 &           0.20 &            0.30 \\
 Deepseek       & noreasoning &    0.20 &    0.20 &    0.10 &    0.20 &    0.20 & famous        &   0.18 &           0.14 &            0.20 &           0.14 &            0.20 \\
 Deepseek       & noreasoning &    0.00 &    0.00 &    0.00 &    0.00 &    0.00 & international &   0.00 &           0.00 &            0.00 &           0.00 &            0.00 \\
 Deepseek       & noreasoning &    0.30 &    0.20 &    0.20 &    0.20 &    0.20 & ribbon        &   0.22 &           0.20 &            0.26 &           0.20 &            0.28 \\
 Deepseek       & reasoning   &    0.30 &    0.00 &    0.20 &    0.20 &    0.10 & campaign      &   0.16 &           0.08 &            0.24 &           0.04 &            0.26 \\
 Deepseek       & reasoning   &    0.00 &    0.30 &    0.00 &    0.00 &    0.00 & famous        &   0.06 &           0.00 &            0.18 &           0.00 &            0.18 \\
 Deepseek       & reasoning   &    0.00 &    0.00 &    0.00 &    0.00 &    0.00 & international &   0.00 &           0.00 &            0.00 &           0.00 &            0.00 \\
 Deepseek       & reasoning   &    0.00 &    0.30 &    0.00 &    0.30 &    0.00 & ribbon        &   0.12 &           0.00 &            0.24 &           0.00 &            0.30 \\
GPT 5          & noreasoning* &    0.20 &    0.20 &    0.20 &    0.20 &    0.20 & campaign      &   0.20 &           0.20 &            0.20 &           0.20 &            0.20 \\
GPT 5          & noreasoning &    0.10 &    0.10 &    0.10 &    0.10 &    0.10 & famous        &   0.10 &           0.10 &            0.10 &           0.10 &            0.10 \\
GPT 5          & noreasoning &    0.00 &    0.00 &    0.00 &    0.00 &    0.00 & international &   0.00 &           0.00 &            0.00 &           0.00 &            0.00 \\
GPT 5          & noreasoning &    0.10 &    0.10 &    0.10 &    0.10 &    0.10 & ribbon        &   0.10 &           0.10 &            0.10 &           0.10 &            0.10 \\
GPT 5          & reasoning**   &    0.10 &    0.20 &    0.20 &    0.20 &    0.20 & campaign      &   0.18 &           0.14 &            0.20 &           0.12 &            0.20 \\
GPT 5          & reasoning   &    0.00 &    0.00 &    0.00 &    0.00 &    0.00 & famous        &   0.00 &           0.00 &            0.00 &           0.00 &            0.00 \\
GPT 5          & reasoning   &    0.00 &    0.00 &    0.00 &    0.00 &    0.00 & international &   0.00 &           0.00 &            0.00 &           0.00 &            0.00 \\
GPT 5          & reasoning   &    0.00 &    0.00 &    0.00 &    0.00 &    0.00 & ribbon        &   0.00 &           0.00 &            0.00 &           0.00 &            0.00 \\
 GPT-oss-120b    & reasoning   &    0.60 &    0.20 &    0.40 &    0.40 &    0.20 & campaign      &   0.36 &           0.24 &            0.48 &           0.24 &            0.52 \\
 GPT-oss-120b    & reasoning   &    0.20 &    0.20 &    0.10 &    0.10 &    0.20 & famous        &   0.16 &           0.12 &            0.20 &           0.10 &            0.20 \\
 GPT-oss-120b    & reasoning   &    0.00 &    0.00 &    0.00 &    0.00 &    0.20 & international &   0.04 &           0.00 &            0.12 &           0.00 &            0.12 \\
 GPT-oss-120b    & reasoning   &    0.70 &    0.20 &    0.20 &    0.20 &    0.20 & ribbon        &   0.30 &           0.20 &            0.50 &           0.20 &            0.60 \\
 GPT-oss-20b     & reasoning   &    0.50 &    0.30 &    0.20 &    0.20 &    0.20 & campaign      &   0.28 &           0.20 &            0.40 &           0.20 &            0.42 \\
 GPT-oss-20b     & reasoning   &    0.20 &    0.20 &    0.20 &    0.00 &    0.20 & famous        &   0.16 &           0.08 &            0.20 &           0.08 &            0.20 \\
 GPT-oss-20b     & reasoning   &    0.00 &    0.00 &    0.00 &    0.00 &    0.00 & international &   0.00 &           0.00 &            0.00 &           0.00 &            0.00 \\
 GPT-oss-20b     & reasoning   &    0.20 &    0.00 &    0.00 &    0.00 &    0.00 & ribbon        &   0.04 &           0.00 &            0.12 &           0.00 &            0.12 \\
 Llama4 Maverick & noreasoning &    0.30 &    0.30 &    0.30 &    0.30 &    0.20 & campaign      &   0.28 &           0.24 &            0.30 &           0.24 &            0.30 \\
 Llama4 Maverick & noreasoning &    0.00 &    0.00 &    0.00 &    0.00 &    0.00 & famous        &   0.00 &           0.00 &            0.00 &           0.00 &            0.00 \\
 Llama4 Maverick & noreasoning &    0.00 &    0.00 &    0.00 &    0.00 &    0.00 & international &   0.00 &           0.00 &            0.00 &           0.00 &            0.00 \\
 Llama4 Maverick & noreasoning &    0.20 &    0.00 &    0.30 &    0.30 &    0.30 & ribbon        &   0.22 &           0.10 &            0.30 &           0.06 &            0.30 \\
 Llama4 Scout    & noreasoning &    0.10 &    0.20 &    0.20 &    0.00 &    0.00 & campaign      &   0.10 &           0.02 &            0.18 &           0.00 &            0.20 \\
 Llama4 Scout    & noreasoning &    0.40 &    0.30 &    0.50 &    0.40 &    0.40 & famous        &   0.40 &           0.34 &            0.46 &           0.34 &            0.46 \\
 Llama4 Scout    & noreasoning &    0.00 &    0.00 &    0.00 &    0.00 &    0.00 & international &   0.00 &           0.00 &            0.00 &           0.00 &            0.00 \\
 Llama4 Scout    & noreasoning &    0.00 &    0.30 &    0.10 &    0.10 &    0.10 & ribbon        &   0.12 &           0.04 &            0.22 &           0.02 &            0.24 \\
 Qwen3 235B          & noreasoning &    0.60 &    0.40 &    0.30 &    0.20 &    0.30 & campaign      &   0.36 &           0.26 &            0.48 &           0.24 &            0.52 \\
 Qwen3 235B          & noreasoning &    0.00 &    0.00 &    0.00 &    0.10 &    0.00 & famous        &   0.02 &           0.00 &            0.06 &           0.00 &            0.08 \\
 Qwen3 235B          & noreasoning &    0.00 &    0.10 &    0.00 &    0.00 &    0.00 & international &   0.02 &           0.00 &            0.06 &           0.00 &            0.08 \\
 Qwen3 235B          & noreasoning &    0.20 &    0.30 &    0.20 &    0.00 &    0.30 & ribbon        &   0.20 &           0.10 &            0.28 &           0.06 &            0.30 \\
 Qwen3 235B          & reasoning   &    0.10 &    0.20 &    0.10 &    0.00 &    0.20 & campaign      &   0.12 &           0.06 &            0.18 &           0.04 &            0.20 \\
 Qwen3 235B          & reasoning   &    0.00 &    0.10 &    0.00 &    0.00 &    0.00 & famous        &   0.02 &           0.00 &            0.06 &           0.00 &            0.08 \\
 Qwen3 235B          & reasoning   &    0.00 &    0.00 &    0.00 &    0.00 &    0.00 & international &   0.00 &           0.00 &            0.00 &           0.00 &            0.00 \\
 Qwen3 235B          & reasoning   &    0.00 &    0.00 &    0.00 &    0.20 &    0.20 & ribbon        &   0.08 &           0.00 &            0.16 &           0.00 &            0.20 \\
 Human          & Human       &    0.00 &    0.10 &  nan    &  nan    &  nan    & campaign      &   0.05 &           0.00 &            0.10 &           0.00 &            0.10 \\
 Human          & Human       &    0.00 &    0.00 &  nan    &  nan    &  nan    & famous        &   0.00 &         nan    &          nan    &         nan    &          nan    \\
 Human          & Human       &    0.00 &    0.00 &  nan    &  nan    &  nan    & international &   0.00 &         nan    &          nan    &         nan    &          nan    \\
 Human          & Human       &    0.00 &    0.00 &  nan    &  nan    &  nan    & ribbon        &   0.00 &         nan    &          nan    &         nan    &          nan    \\
\bottomrule
\end{tabular}
\end{adjustbox}

\caption*{\footnotesize \textit{Note.} * Reasoning effort: minimal. This condition refers to cases in which the model applies little to no reasoning. GPT-5, in particular, does not currently allow users to explicitly disable reasoning. ** Reasoning effort: high. }
\end{table}

\begin{table}[htbp]
\centering
\caption{Populism Score: Turkey – Recep Tayyip Erdoğan}
\label{tab:populism_scores_tk}
\begin{adjustbox}{max width=\textwidth}
\begin{tabular}{llrrrrrlrrrrr}
\toprule
\textbf{Model} & \textbf{Type} & \textbf{Run 1} & \textbf{Run 2} & \textbf{Run 3} & \textbf{Run 4} & \textbf{Run 5} & \textbf{Speech} & \textbf{Mean} & \makecell{\textbf{95\% CI}\\\textbf{Low}} & \makecell{\textbf{95\% CI}\\\textbf{High}} & \makecell{\textbf{99\% CI}\\\textbf{Low}} & \makecell{\textbf{99\% CI}\\\textbf{High}} \\
\midrule
 Deepseek       & noreasoning &    1.60 &    1.70 &    1.70 &    1.60 &    1.70 & campaign      &   1.66 &           1.62 &            1.70 &           1.60 &            1.70 \\
 Deepseek       & noreasoning &    1.80 &    1.80 &    2.00 &    1.40 &    1.80 & famous        &   1.76 &           1.56 &            1.92 &           1.48 &            1.92 \\
 Deepseek       & noreasoning &    0.30 &    0.20 &    0.30 &    0.40 &    0.70 & international &   0.38 &           0.26 &            0.56 &           0.24 &            0.60 \\
 Deepseek       & noreasoning &    1.40 &    1.40 &    1.20 &    1.80 &    1.50 & ribbon        &   1.46 &           1.30 &            1.64 &           1.28 &            1.72 \\
 Deepseek       & reasoning   &    1.60 &    1.50 &    1.80 &    1.90 &    1.80 & campaign      &   1.72 &           1.60 &            1.84 &           1.56 &            1.86 \\
 Deepseek       & reasoning   &    0.90 &    2.00 &    1.70 &    1.70 &    1.90 & famous        &   1.64 &           1.28 &            1.90 &           1.22 &            1.96 \\
 Deepseek       & reasoning   &    0.00 &    0.10 &    0.70 &    0.00 &    0.30 & international &   0.22 &           0.02 &            0.48 &           0.00 &            0.56 \\
 Deepseek       & reasoning   &    1.50 &    1.10 &    2.00 &    1.70 &    1.50 & ribbon        &   1.56 &           1.30 &            1.82 &           1.26 &            1.90 \\
GPT 5          & noreasoning* &    1.40 &    1.40 &    1.60 &    1.50 &    1.40 & campaign      &   1.46 &           1.40 &            1.54 &           1.40 &            1.54 \\
GPT 5          & noreasoning &    1.70 &    1.60 &    1.60 &    1.60 &    1.60 & famous        &   1.62 &           1.60 &            1.66 &           1.60 &            1.68 \\
GPT 5          & noreasoning &    0.70 &    0.60 &    0.60 &    0.60 &    0.40 & international &   0.58 &           0.48 &            0.66 &           0.44 &            0.68 \\
GPT 5          & noreasoning &    1.30 &    1.20 &    1.20 &    1.20 &    1.40 & ribbon        &   1.26 &           1.20 &            1.34 &           1.20 &            1.36 \\
GPT 5          & reasoning**   &    1.70 &    1.60 &    1.80 &    1.70 &    1.70 & campaign      &   1.70 &           1.64 &            1.76 &           1.64 &            1.78 \\
GPT 5          & reasoning   &    1.60 &    1.10 &    1.70 &    1.80 &    1.70 & famous        &   1.58 &           1.34 &            1.74 &           1.22 &            1.76 \\
GPT 5          & reasoning   &    0.20 &    0.20 &    0.10 &    0.20 &    0.20 & international &   0.18 &           0.14 &            0.20 &           0.12 &            0.20 \\
GPT 5          & reasoning   &    1.30 &    1.50 &    1.40 &    1.40 &    1.60 & ribbon        &   1.44 &           1.36 &            1.54 &           1.34 &            1.56 \\
 GPT-oss-120b    & reasoning   &    1.70 &    1.70 &    1.50 &    1.90 &    1.60 & campaign      &   1.68 &           1.58 &            1.80 &           1.56 &            1.82 \\
 GPT-oss-120b    & reasoning   &    1.70 &    1.70 &    1.90 &    1.70 &    1.70 & famous        &   1.74 &           1.70 &            1.82 &           1.70 &            1.86 \\
 GPT-oss-120b    & reasoning   &    0.80 &    0.60 &    0.00 &    0.20 &    0.30 & international &   0.38 &           0.12 &            0.62 &           0.08 &            0.70 \\
 GPT-oss-120b    & reasoning   &    1.70 &    1.90 &    1.50 &    1.80 &    1.60 & ribbon        &   1.70 &           1.58 &            1.82 &           1.56 &            1.86 \\
 GPT-oss-20b     & reasoning   &    0.40 &    1.40 &    0.20 &    0.40 &    0.60 & campaign      &   0.60 &           0.32 &            1.04 &           0.28 &            1.20 \\
 GPT-oss-20b     & reasoning   &    1.80 &    1.70 &    1.80 &    1.50 &    1.30 & famous        &   1.62 &           1.44 &            1.78 &           1.38 &            1.80 \\
 GPT-oss-20b     & reasoning   &    0.80 &    0.00 &    0.30 &    0.20 &    0.40 & international &   0.34 &           0.14 &            0.58 &           0.10 &            0.64 \\
 GPT-oss-20b     & reasoning   &    0.80 &    1.10 &    0.10 &    0.40 &    0.70 & ribbon        &   0.62 &           0.30 &            0.92 &           0.22 &            1.02 \\
 Llama4 Maverick & noreasoning &    0.40 &    0.40 &    0.40 &    0.20 &    0.30 & campaign      &   0.34 &           0.26 &            0.40 &           0.24 &            0.40 \\
 Llama4 Maverick & noreasoning &    1.70 &    1.70 &    1.70 &    1.70 &    1.70 & famous        &   1.70 &           1.70 &            1.70 &           1.70 &            1.70 \\
 Llama4 Maverick & noreasoning &    0.00 &    0.00 &    0.00 &    0.20 &    0.10 & international &   0.06 &           0.00 &            0.14 &           0.00 &            0.16 \\
 Llama4 Maverick & noreasoning &    1.30 &    1.50 &    1.00 &    1.00 &    1.00 & ribbon        &   1.16 &           1.00 &            1.32 &           1.00 &            1.40 \\
 Llama4 Scout    & noreasoning &    0.40 &    0.10 &    0.10 &    0.10 &    0.30 & campaign      &   0.20 &           0.10 &            0.32 &           0.10 &            0.36 \\
 Llama4 Scout    & noreasoning &    1.70 &    1.70 &    1.70 &    1.70 &    0.90 & famous        &   1.54 &           1.22 &            1.70 &           1.22 &            1.70 \\
 Llama4 Scout    & noreasoning &    0.00 &    0.00 &    0.00 &    0.00 &    0.10 & international &   0.02 &           0.00 &            0.06 &           0.00 &            0.06 \\
 Llama4 Scout    & noreasoning &    0.30 &    0.40 &    0.40 &    0.40 &    0.80 & ribbon        &   0.46 &           0.34 &            0.64 &           0.34 &            0.70 \\
 Qwen3 235B          & noreasoning &    1.60 &    1.70 &    1.20 &    1.50 &    1.60 & campaign      &   1.52 &           1.34 &            1.64 &           1.30 &            1.66 \\
 Qwen3 235B          & noreasoning &    1.40 &    1.70 &    1.80 &    1.40 &    1.80 & famous        &   1.62 &           1.46 &            1.78 &           1.40 &            1.80 \\
 Qwen3 235B          & noreasoning &    0.00 &    0.20 &    0.20 &    0.40 &    0.30 & international &   0.22 &           0.10 &            0.34 &           0.08 &            0.36 \\
 Qwen3 235B          & noreasoning &    1.40 &    1.60 &    1.70 &    1.60 &    1.80 & ribbon        &   1.62 &           1.48 &            1.72 &           1.46 &            1.76 \\
 Qwen3 235B          & reasoning   &    1.50 &    1.80 &    1.50 &    1.90 &    1.60 & campaign      &   1.66 &           1.52 &            1.80 &           1.50 &            1.84 \\
 Qwen3 235B          & reasoning   &    2.00 &    2.00 &    2.00 &    1.90 &    1.80 & famous        &   1.94 &           1.86 &            2.00 &           1.84 &            2.00 \\
 Qwen3 235B          & reasoning   &    0.10 &    0.10 &    0.10 &    0.00 &    0.10 & international &   0.08 &           0.04 &            0.10 &           0.04 &            0.10 \\
 Qwen3 235B          & reasoning   &    1.60 &    1.60 &    1.60 &    1.60 &    1.60 & ribbon        &   1.60 &           1.60 &            1.60 &           1.60 &            1.60 \\
 Human          & Human       &    2.00 &    2.00 &  nan    &  nan    &  nan    & campaign      &   2.00 &         nan    &          nan    &         nan    &          nan    \\
 Human          & Human       &    1.80 &    2.00 &  nan    &  nan    &  nan    & famous        &   1.90 &           1.80 &            2.00 &         nan    &          nan    \\
 Human          & Human       &    0.00 &    0.40 &  nan    &  nan    &  nan    & international &   0.20 &           0.00 &            0.40 &           0.00 &            0.40 \\
 Human          & Human       &    1.60 &    2.00 &  nan    &  nan    &  nan    & ribbon        &   1.70 &           1.60 &            2.00 &         nan    &          nan    \\
\bottomrule
\end{tabular}
\end{adjustbox}
\caption*{\footnotesize \textit{Note.} * Reasoning effort: minimal. This condition refers to cases in which the model applies little to no reasoning. GPT-5, in particular, does not currently allow users to explicitly disable reasoning. ** Reasoning effort: high. }
\end{table}

\begin{table}[htbp]
\centering
\caption{Populism Score: Montenegro – Milo Đukanović}
\label{tab:populism_scores_me}
\begin{adjustbox}{max width=\textwidth}
\begin{tabular}{llrrrrrlrrrrr}
\toprule
\textbf{Model} & \textbf{Type} & \textbf{Run 1} & \textbf{Run 2} & \textbf{Run 3} & \textbf{Run 4} & \textbf{Run 5} & \textbf{Speech} & \textbf{Mean} & \makecell{\textbf{95\% CI}\\\textbf{Low}} & \makecell{\textbf{95\% CI}\\\textbf{High}} & \makecell{\textbf{99\% CI}\\\textbf{Low}} & \makecell{\textbf{99\% CI}\\\textbf{High}} \\
\midrule
 Deepseek       & noreasoning &    0.80 &    0.70 &    0.40 &    0.30 &    0.80 & campaign      &   0.60 &             0.42 &              0.78 &             0.38 &              0.80 \\
 Deepseek       & noreasoning &    0.20 &    0.30 &    0.30 &    0.30 &    0.20 & famous        &   0.26 &             0.22 &              0.30 &             0.20 &              0.30 \\
 Deepseek       & noreasoning &    0.00 &    0.00 &    0.00 &    0.10 &    0.20 & international &   0.06 &             0.00 &              0.14 &             0.00 &              0.16 \\
 Deepseek       & noreasoning &    0.00 &    0.00 &    0.00 &    0.00 &    0.00 & ribbon        &   0.00 &             0.00 &              0.00 &             0.00 &              0.00 \\
 Deepseek       & reasoning   &    0.30 &    0.00 &    0.20 &    0.30 &    0.00 & campaign      &   0.16 &             0.04 &              0.28 &             0.00 &              0.30 \\
 Deepseek       & reasoning   &    0.00 &    0.00 &    0.00 &    0.20 &    0.00 & famous        &   0.04 &             0.00 &              0.12 &             0.00 &              0.16 \\
 Deepseek       & reasoning   &    0.00 &    0.00 &    0.00 &    0.00 &    0.00 & international &   0.00 &             0.00 &              0.00 &             0.00 &              0.00 \\
 Deepseek       & reasoning   &    0.00 &    0.00 &    0.00 &    0.00 &    0.00 & ribbon        &   0.00 &             0.00 &              0.00 &             0.00 &              0.00 \\
 GPT-oss-120b    & reasoning   &    0.70 &    0.20 &    0.30 &    0.70 &    0.20 & campaign      &   0.42 &             0.22 &              0.62 &             0.20 &              0.70 \\
 GPT-oss-120b    & reasoning   &    0.20 &    0.20 &    0.20 &    0.20 &    0.70 & famous        &   0.30 &             0.20 &              0.50 &             0.20 &              0.50 \\
 GPT-oss-120b    & reasoning   &    0.00 &    0.20 &    0.10 &    0.00 &    0.30 & international &   0.12 &             0.02 &              0.22 &             0.00 &              0.24 \\
 GPT-oss-120b    & reasoning   &    0.30 &    0.10 &    0.20 &    0.00 &    0.20 & ribbon        &   0.16 &             0.06 &              0.24 &             0.04 &              0.26 \\
 GPT-oss-20b     & reasoning   &    0.20 &    0.00 &    0.20 &    0.20 &    0.20 & campaign      &   0.16 &             0.08 &              0.20 &             0.04 &              0.20 \\
 GPT-oss-20b     & reasoning   &    0.20 &    0.20 &    0.20 &    0.00 &    0.20 & famous        &   0.16 &             0.08 &              0.20 &             0.08 &              0.20 \\
 GPT-oss-20b     & reasoning   &    0.00 &    0.00 &    0.00 &    0.20 &    0.00 & international &   0.04 &             0.00 &              0.12 &             0.00 &              0.16 \\
 GPT-oss-20b     & reasoning   &    0.00 &    0.00 &    0.00 &    0.00 &    0.00 & ribbon        &   0.00 &             0.00 &              0.00 &             0.00 &              0.00 \\
 Llama4 Maverick & noreasoning &    0.20 &    0.00 &    0.00 &    1.00 &    0.00 & campaign      &   0.24 &             0.00 &              0.64 &             0.00 &              0.80 \\
 Llama4 Maverick & noreasoning &    0.00 &    0.30 &    0.00 &    0.00 &    0.00 & famous        &   0.06 &             0.00 &              0.18 &             0.00 &              0.24 \\
 Llama4 Maverick & noreasoning &    0.00 &    0.00 &    0.00 &    0.00 &    0.00 & international &   0.00 &             0.00 &              0.00 &             0.00 &              0.00 \\
 Llama4 Maverick & noreasoning &    0.00 &    0.00 &    0.00 &    0.00 &    0.00 & ribbon        &   0.00 &             0.00 &              0.00 &             0.00 &              0.00 \\
 Llama4 Scout    & noreasoning &    0.10 &    0.50 &    0.20 &    0.30 &    0.50 & campaign      &   0.32 &             0.20 &              0.44 &             0.16 &              0.50 \\
 Llama4 Scout    & noreasoning &    0.30 &    0.30 &    0.40 &    0.10 &    0.20 & famous        &   0.25 &             0.12 &              0.35 &             0.10 &              0.38 \\
 Llama4 Scout    & noreasoning &    0.00 &    0.00 &    0.00 &    0.00 &    0.00 & international &   0.00 &             0.00 &              0.00 &             0.00 &              0.00 \\
 Llama4 Scout    & noreasoning &    0.00 &    0.30 &    0.20 &    0.20 &    0.00 & ribbon        &   0.14 &             0.04 &              0.24 &             0.00 &              0.26 \\
 Qwen3 235B          & noreasoning &    0.30 &    0.20 &    0.20 &    0.20 &    0.40 & campaign      &   0.26 &             0.20 &              0.32 &             0.20 &              0.36 \\
 Qwen3 235B          & noreasoning &    0.00 &    0.30 &    0.00 &    0.00 &    0.00 & famous        &   0.06 &             0.00 &              0.18 &             0.00 &              0.24 \\
 Qwen3 235B          & noreasoning &    0.00 &    0.00 &    0.00 &    0.00 &    0.00 & international &   0.00 &             0.00 &              0.00 &             0.00 &              0.00 \\
 Qwen3 235B          & noreasoning &    0.00 &    0.00 &    0.00 &    0.10 &    0.00 & ribbon        &   0.02 &             0.00 &              0.06 &             0.00 &              0.08 \\
 Qwen3 235B          & reasoning   &    0.20 &    0.20 &    0.00 &    1.10 &    0.10 & campaign      &   0.32 &             0.08 &              0.72 &             0.04 &              0.74 \\
 Qwen3 235B          & reasoning   &    0.20 &    0.00 &    0.20 &    0.00 &    0.00 & famous        &   0.08 &             0.00 &              0.16 &             0.00 &              0.16 \\
 Qwen3 235B          & reasoning   &    0.00 &    0.00 &    0.00 &    0.00 &    0.00 & international &   0.00 &             0.00 &              0.00 &             0.00 &              0.00 \\
 Qwen3 235B          & reasoning   &    0.00 &    0.00 &    0.00 &    0.00 &    0.00 & ribbon        &   0.00 &             0.00 &              0.00 &             0.00 &              0.00 \\
 Human          & Human       &    0.00 &    0.00 &  nan    &  nan    &  nan    & campaign      &   0.00 &           nan    &            nan    &           nan    &            nan    \\
 Human          & Human       &    0.00 &    1.00 &  nan    &  nan    &  nan    & famous        &   0.50 &             0.10 &              0.10 &           nan    &            nan    \\
 Human          & Human       &    0.00 &    0.00 &  nan    &  nan    &  nan    & international &   0.00 &           nan    &            nan    &           nan    &            nan    \\
 Human          & Human       &    0.00 &    0.00 &  nan    &  nan    &  nan    & ribbon        &   0.00 &           nan    &            nan    &           nan    &            nan    \\
GPT 5          & noreasoning* &    0.60 &    0.60 &    0.60 &    0.50 &    0.60 & campaign      &   0.58 &             0.54 &              0.60 &             0.52 &              0.60 \\
GPT 5          & noreasoning &    0.30 &    0.20 &    0.30 &    0.20 &    0.30 & famous        &   0.26 &             0.22 &              0.30 &             0.20 &              0.30 \\
GPT 5          & noreasoning &    0.00 &    0.10 &    0.00 &    0.10 &    0.10 & international &   0.06 &             0.02 &              0.10 &             0.00 &              0.10 \\
GPT 5          & noreasoning &    0.00 &    0.00 &    0.00 &    0.00 &    0.00 & ribbon        &   0.00 &             0.00 &              0.00 &             0.00 &              0.00 \\
GPT 5          & reasoning**   &    0.10 &    0.20 &    0.10 &    0.20 &    0.20 & campaign      &   0.16 &             0.12 &              0.20 &             0.10 &              0.20 \\
GPT 5          & reasoning   &    0.00 &    0.00 &    0.00 &    0.00 &    0.00 & famous        &   0.00 &             0.00 &              0.00 &             0.00 &              0.00 \\
GPT 5          & reasoning   &    0.00 &    0.00 &    0.00 &    0.00 &    0.00 & international &   0.00 &             0.00 &              0.00 &             0.00 &              0.00 \\
GPT 5          & reasoning   &    0.00 &    0.00 &    0.00 &    0.00 &    0.00 & ribbon        &   0.00 &             0.00 &              0.00 &             0.00 &              0.00 \\
\bottomrule
\end{tabular}
\end{adjustbox}
\caption*{\footnotesize \textit{Note.} * Reasoning effort: minimal. This condition refers to cases in which the model applies little to no reasoning. GPT-5, in particular, does not currently allow users to explicitly disable reasoning. ** Reasoning effort: high. }

\end{table}


\newpage

\section{Error and Calibration}\label{app:error-calibration}

Having established strong agreement between (top) AI systems and human graders, we now examine the precision and calibration of these models. We track three error statistics and two calibration diagnostics to assess both the magnitude of AI-human differences and whether those differences follow systematic patterns that could undermine practical application. 

\subsection{Error Metrics}\label{app:error}
We begin with Mean Absolute Error (MAE), which measures the average size of AI–Human disagreements regardless of direction. MAE simply averages |AI – Human| across all items, providing a direct measure of typical scoring differences in the original 0-2 scale units. In practice, it ignores direction: being +0.10 or -0.10 off both count as 0.10. This means that an MAE = 0.10 would indicate that the AI is off by about a tenth of a point on average (approximately 5\% of the full scale range), while MAE = 0.25 represents larger but still moderate disagreements of about one-quarter point (12.5\% of range).

Next, we complement MAE with the Root Mean Square Error (RMSE), which provides additional sensitivity to large disagreements by squaring each error before averaging and taking the square root. While RMSE shares MAE’s original scale units, it penalizes large misses more heavily – making it particularly useful for detecting whether AI systems occasionally produce dramatic scoring failures. RMSE will always equal or exceed MAE, with a larger RMSE relative to MAE indicating the presence of substantial outlier disagreements rather than consistent small errors. 

Finally, we close the error diagnostics by examining signed bias (mean AI – Human difference), which reveals systematic directional errors that averaging masks. Negative bias indicates that AI systems consistently under-score populism relative to human graders, while positive bias suggests over-scoring tendencies. Together, these three metrics – MAE for typical error size, RMSE for sensitivity to large disagreements, and signed bias for directional patterns – provide a comprehensive picture of AI scoring error patterns. 

\subsection{Calibration Diagnostics}\label{app:calibration}

Moving to calibration, we start by assessing whether AI systems not only 
agree with humans but also do so with proper scaling and level. We use simple 
linear calibration regression by regressing human scores on AI scores:
\[
\text{Human} = \beta_{0} + \beta_{1} \cdot \text{AI}.
\]
This helps us understand if disagreements between AI and human graders 
are a product of mis-calibration. 

The intercept ($\beta_{0}$) assesses level calibration, or whether AI 
models show consistent directional bias across the measurement range. 
Perfect level calibration requires $\beta_{0} \approx 0$; when the 
95\% confidence interval excludes zero, the model exhibits systematic 
offset (consistently scoring higher or lower than humans, regardless of 
speech content). The slope ($\beta_{1}$) captures scale calibration, or 
whether AI systems appropriately differentiate between low and high 
populism speeches. In this specification (Human on AI),the perfect scale 
calibration requires $\beta_{1} \approx 1$; $\beta_{1} > 1$ indicates AI 
scale compression (AI under-reacts, pulling extremes toward the center), 
while $\beta_{1} < 1$ indicates expansion (AI over-reacts, exaggerating 
differences).

We conclude with Bland-Altman (BA) analysis, which addresses a key 
question: Are AI--human disagreements consistent across the populism 
scale, or do errors systematically vary with score level? BA plots the 
difference $(\text{AI} - \text{Human})$ against the pairwise mean 
$\big( (\text{AI} + \text{Human}) / 2 \big)$, revealing whether 
disagreements concentrate at particular scale regions---for instance, 
whether AI systems struggle more with moderate populism scores than with 
clear high/low cases.

Two key statistics summarize the BA pattern: the average bias (mean 
AI $-$ Human difference) and the limits of agreement (LoA), calculated 
as:
\[
\text{LoA} = \text{bias} \pm 1.96 \times \text{SD(differences)}.
\]
Under assumptions of roughly normal and independent errors, approximately 
95\% of AI--human differences should fall within these limits. Narrow 
LoA indicates consistent disagreement magnitude across the scale, while wide LoA suggests that the error size varies substantially depending on the Specific speech is being coded. While our previous metrics all summarize performance across runs (i.e., they are model-level metrics), we examine BA at the run-level to detect systematic calibration patterns that might be masked by across-run averaging.

Below, Table~4 presents the MAE, RMSE, and signed biases for each model, while Table~5 reports the mean intercept, mean slope, and mean $R^{2}$ for the simple linear calibration regressions. Figure~2 depicts BA plots for the best runs of the two best models and for the best run of the weakest model.

\begin{table}[htbp]
\footnotesize
\centering
\caption{Error Metrics by Model (MAE, RMSE, Bias)}
\label{tab:error_metrics}
\begin{tabular}{@{}llccc@{}}
\toprule
\textbf{Model} & \textbf{Architecture} & \textbf{MAE} & \textbf{RMSE} & \textbf{Bias} \\ \midrule
GPT-5 & Reasoning (high) & 0.146 \newline (0.027) & 0.233 \newline (0.050) & -0.098 \newline (0.028) \\
GPT-5 & Standard (minimal) & 0.245 \newline (0.016) & 0.320 \newline (0.017) & -0.016 \newline (0.011) \\
GPT-oss 120B & MoE Reasoning & 0.221 \newline (0.061) & 0.272 \newline (0.079) & +0.075 \newline (0.066) \\
GPT-oss 20B & MoE Reasoning & 0.335 \newline (0.076) & 0.559 \newline (0.159) & -0.200 \newline (0.078) \\
DeepSeek R1 & MoE Reasoning & 0.169 \newline (0.058) & 0.257 \newline (0.086) & -0.061 \newline (0.067) \\
DeepSeek V3 & MoE Standard & 0.209 \newline (0.037) & 0.280 \newline (0.044) & +0.034 \newline (0.040) \\
Qwen3 235B & MoE Reasoning (high) & 0.133 \newline (0.038) & 0.223 \newline (0.079) & -0.043 \newline (0.040) \\
Qwen3 235B & MoE Standard & 0.192 \newline (0.038) & 0.270 \newline (0.049) & -0.041 \newline (0.044) \\
Llama 4 Maverick & MoE Standard & 0.311 \newline (0.062) & 0.556 \newline (0.066) & -0.196 \newline (0.037) \\
Llama 4 Scout & MoE Standard & 0.413 \newline (0.038) & 0.684 \newline (0.030) & -0.238 \newline (0.031) \\ \bottomrule
\end{tabular}
\end{table}

\begin{table}[htbp]
\footnotesize
\centering
\caption{Calibration Regression Summaries by Model (Intercept, Slope, $R^2$)}
\label{tab:calibration_regression}
\begin{tabular}{@{}llcccc@{}}
\toprule
\textbf{Model} & \textbf{Architecture} & \textbf{Runs} & \textbf{Intercept} & \textbf{Slope} & \textbf{$R^2$} \\ \midrule
\multirow{2}{*}{GPT-5} & \multirow{2}{*}{Reasoning (high)} & \multirow{2}{*}{5} 
 & 0.0261568 & 1.168067 & 0.9479944 \\
 &  &  & (0.0113223) & (0.0716100) & (0.0204445) \\[0.3em]

\multirow{2}{*}{GPT-5} & \multirow{2}{*}{Standard (minimal)} & \multirow{2}{*}{5} 
 & -0.1429972 & 1.308030 & 0.8883976 \\
 &  &  & (0.0049944) & (0.0328442) & (0.0195649) \\[0.3em]

\multirow{2}{*}{GPT-oss 120B} & \multirow{2}{*}{MoE Reasoning} & \multirow{2}{*}{5} 
 & -0.1683036 & 1.153328 & 0.9080602 \\
 &  &  & (0.0863175) & (0.0833291) & (0.0666842) \\[0.3em]

\multirow{2}{*}{GPT-oss 20B} & \multirow{2}{*}{MoE Reasoning} & \multirow{2}{*}{5} 
 & 0.0925538 & 1.328426 & 0.5995936 \\
 &  &  & (0.1342727) & (0.3551010) & (0.2473296) \\[0.3em]

\multirow{2}{*}{DeepSeek R1} & \multirow{2}{*}{MoE Reasoning} & \multirow{2}{*}{5} 
 & 0.0197423 & 1.098702 & 0.9089600 \\
 &  &  & (0.0343635) & (0.1034719) & (0.0489237) \\[0.3em]

\multirow{2}{*}{DeepSeek V3} & \multirow{2}{*}{MoE Standard} & \multirow{2}{*}{5} 
 & -0.1307498 & 1.171742 & 0.8974385 \\
 &  &  & (0.0410053) & (0.0338172) & (0.0380681) \\[0.3em]

\multirow{2}{*}{Qwen3 235B} & \multirow{2}{*}{MoE Reasoning (high)} & \multirow{2}{*}{5} 
 & 0.0252999 & 1.041053 & 0.9240385 \\
 &  &  & (0.0098243) & (0.0774916) & (0.0689529) \\[0.3em]

\multirow{2}{*}{Qwen3 235B} & \multirow{2}{*}{MoE Standard} & \multirow{2}{*}{5} 
 & -0.0421470 & 1.172454 & 0.9084351 \\
 &  &  & (0.0397188) & (0.0678548) & (0.0363577) \\[0.3em]

\multirow{2}{*}{Llama 4 Maverick} & \multirow{2}{*}{MoE Standard} & \multirow{2}{*}{5} 
 & 0.1526475 & 1.140714 & 0.5854542 \\
 &  &  & (0.0231706) & (0.1425217) & (0.1367799) \\[0.3em]

\multirow{2}{*}{Llama 4 Scout} & \multirow{2}{*}{MoE Standard} & \multirow{2}{*}{5} 
 & 0.1840526 & 1.200254 & 0.3763293 \\
 &  &  & (0.0964273) & (0.3978559) & (0.1114074) \\ \bottomrule
\end{tabular}
\end{table}

\begin{figure}[H]  
    \centering
    \includegraphics[width=\textwidth]{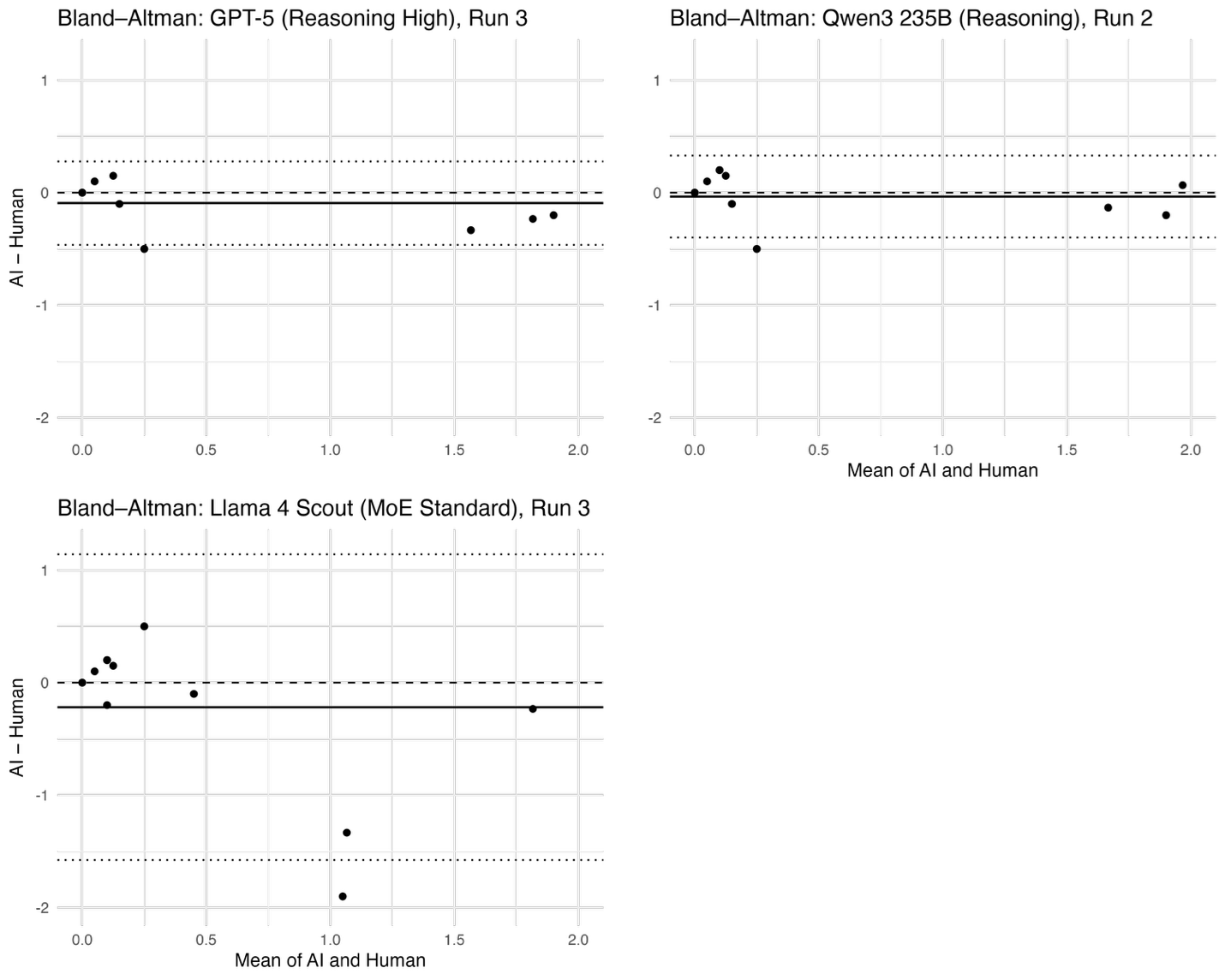}
    \caption{\footnotesize Bland--Altman plots (best runs): GPT-5 Reasoning (run 3), Qwen3 235B Reasoning (run 2), and Llama~4 Scout Standard (run 3).}
    \caption*{\footnotesize \textit{Note.} Bland--Altman plots for the best runs of the two best models (GPT-5 Reasoning high, run 3; Qwen3 235B Reasoning, run 2) and for the best run of the weakest model (Llama~4 Scout Standard, run 3). Traced lines indicate mean bias; dotted lines show the upper and lower limits of agreement (LoA).}
    \label{fig:llm_scores}
\end{figure}

In line with what we identified when analyzing the AI vs.\ Human agreement, run-aggregated error summaries reveal substantial performance differences across models, with the strongest systems achieving precise agreement while weaker models show marked deviations from human scoring. Qwen3 (reasoning) demonstrates the smallest average errors with MAE $= 0.133$ and RMSE $= 0.223$, accompanied by minimal negative bias ($-0.043$)---indicating it scores slightly below humans on average. GPT-5 (reasoning) performs similarly in terms of error magnitude (MAE $= 0.146$, RMSE $= 0.233$) but exhibits a more pronounced negative bias ($-0.098$), suggesting a stronger tendency toward conservative scoring. In both cases, RMSE is $\sim 60$--$70\%$ larger than MAE (ratios $\approx 1.6$--$1.7$), indicating occasional larger disagreements despite small typical errors on the $0$--$2$ scale. Thus, misses larger than the average are present but not dominant, rather than ``rare.''

In sharp contrast, weaker open-source models demonstrate substantially larger errors across all speeches. Llama~4 Scout posts MAE $= 0.413$ and RMSE $= 0.684$ with severe negative bias ($-0.238$), while GPT-oss 20B shows MAE $= 0.335$, RMSE $= 0.559$, and bias $= -0.200$. These error magnitudes represent a qualitative difference in coding reliability: where top models miss by roughly one-tenth of a point (approximately $6$--$7\%$ of the full scale), weaker models deviate by one-third to nearly half a point ($17$--$21\%$ of the scale)---errors large enough to systematically misclassify speeches across populism categories. The pronounced negative biases across all weaker models (with the exception of GPT-oss 120B and DeepSeek V3) indicate systematic underscoring that compounds their precision problems. Practically, this means weaker models fail the fundamental requirement for reliable content analysis: they cannot consistently approximate human judgment within acceptable margins of disagreement.

Calibration-wise, as Table \ref{tab:calibration_regression} shows, Qwen3 (reasoning) demonstrates near-optimal calibration with an average $\beta_{1} = 1.041$ (SD $= 0.077$), $\beta_{0} = 0.025$ (SD $= 0.010$), and $R^{2} = 0.924$---the slope closest to $1$ indicates minimal scale distortion, while the near-zero intercept confirms absence of systematic level bias. GPT-5 (reasoning high) shows comparable metrics, though with somewhat greater miscalibration and scale compression (average $\beta_{1} = 1.168$ (SD $= 0.072$), $\beta_{0} = 0.026$ (SD $= 0.011$), and $R^{2} = 0.948$).

More generally, slopes consistently above 1.0 across top models reveal a systematic pattern: relative to human HG, AI systems compress the scale – slightly over-scoring near the bottom (reluctant to assign true minimum) and under-scoring near the top (hesitant to assign maxima). Substantively, this suggests a “cautious” scoring strategy: models are reluctant to assign extreme values in either direction, preferring to hedge toward the scale center even when textual evidence might strongly support anchor positions.

Finally, we turn to Bland–Altman analysis. Because BA is a run-level diagnostic, we plot only the best run for the two best-performing models (GPT-5 Reasoning High and Qwen3 Reasoning) and for the best run of the weakest model (Llama 4 Scout). The plots in Figure 2 reveal where and how calibration breaks down across the populism scale. Both GPT-5 and Qwen3 (reasoning) demonstrate similar patterns: consistent negative bias (mean bias, the traced lines below zero) with limits of agreement (LoA; upper and lower dotted lines) spanning approximately -0.5 to +0.3. The relatively narrow band between these dotted lines indicates that even for the best models, individual speech errors typically stay within half a point, with most data points clustered much closer to the mean bias line. Crucially, both models show stable error variance across the scale – the vertical scatter of points remains roughly constant across different score ranges, meaning disagreements do not systematically worsen at particular parts of the populism scale. 

Conversely, Llama 4 Scout reveals the hallmarks of poor calibration: substantially larger negative bias and much wider limits of agreement (approximately -1.8 to +1.0). The pronounced scatter and systematic displacement confirm that weaker models not only miss the correct level but also fail to maintain consistent error patterns.

Overall, the error and calibration diagnostics reinforce our core findings: the best reasoning models demonstrate systematic, predictable behavior suitable for automated holistic grading. While top models exhibit conservative tendencies via scale compression and negative bias, these patterns are consistent and potentially correctable – reflecting a stable scoring philosophy rather than random error. By contrast, weaker models show larger errors and poor calibration properties that render them unreliable for research applications. Researchers using GPT-5 and Qwen3 should anticipate modest under-scoring of highly populist texts.

\subsection{Leaderboard}\label{app:leader}

Because our objective is to test whether AI systems can stand in for 
human graders in SHG, we close by ranking models using a synthesis of 
The evidence presented so far is rather than a single statistic. We 
prioritize interchangeability (ICC, CCC), error magnitude (MAE, RMSE), 
calibration quality from \(\text{Human} = \beta_{0} + \beta_{1} \cdot \text{AI}\) (closeness of $\beta_{1}$ to $1$, $\beta_{0}$ to $0$, and $R^{2}$), 
pooled reliability with humans (Krippendorff’s $\alpha$), and 
Bland--Altman patterns (limits of agreement and stability across the 
scale). Ties are broken by calibration closeness and BA tightness.

In the first place, the best overall model is Qwen3 235B (Reasoning). 
It posts the lowest errors (MAE $= 0.133$, RMSE $= 0.223$) with a small 
negative bias ($-0.043$). Calibration is closest to ideal 
($\beta_{1} = 1.041$, $\beta_{0} \approx 0.025$, $R^{2} \approx 0.924$), 
and agreement is excellent (ICC $= .958$, CCC $= .954$). Pooled 
reliability with humans is very high (Krippendorff’s $\alpha = .957$; 
AI-only $\alpha = .958$). The BA plot shows a tight, stable band, 
consistent with mild scale compression and modest under-scoring.

Right behind it, we have GPT-5 (Reasoning). Errors remain low 
(MAE $= 0.146$, RMSE $= 0.233$) with a larger negative bias ($-0.098$). 
Agreement is likewise very strong (ICC $= .951$, CCC $= 0.947$) and 
pooled reliability is the best overall ($\alpha = .969$; AI-only 
$\alpha = .981$), but calibration shows more compression 
($\beta_{1} = 1.168$, $\beta_{0} \approx 0.026$, $R^{2} \approx 0.948$). 
The BA pattern is comparably tight, indicating stable speech-level 
disagreements.

At the other end of the ranking, next-to-last is GPT-oss 20B (MoE 
Reasoning). Errors are high (MAE $= 0.335$, RMSE $= 0.559$) with marked 
under-scoring (bias $= -0.200$), middling agreement (ICC $= .644$, 
CCC $= .626$), borderline pooled reliability ($\alpha = .714$; AI-only 
$\alpha = .797$) and distorted scaling ($\beta_{1} = 1.328$, 
$\beta_{0} \approx 0.093$, $R^{2} \approx 0.600$). The BA plot likewise 
shows a wide band, consistent with noisy calibration.

Finally, at the end of the ranking, last place is Llama~4 Scout 
(Standard). Errors are the largest (MAE $= 0.413$, RMSE $= 0.684$) with 
substantial under-scoring (bias $= -0.238$), weak agreement (ICC $= .482$, 
CCC $= .460$), low pooled reliability ($\alpha = .646$; AI-only 
$\alpha = .872$), and poor calibration ($\beta_{1} = 1.200$, 
$\beta_{0} \approx 0.184$, $R^{2} \approx 0.376$). The BA limits are 
wide ($\approx -1.8$ to $+1.0$) with scattered residuals.

Taken together, our results show that automated SHG is feasible with the top reasoning models. Qwen3 235B (open-weight) and GPT-5 achieve near-interchangeability with human graders, with modest negative bias, and mild scale compression. The fact that an open-weight model matches a flagship proprietary system is another intriguing finding, as it matters for auditability, cost, and reproducibility, while it also indicates that researchers can attain human-comparable scoring without relying on closed APIs. By contrast, models with lower effective capacity – GPT-oss 20B and Llama 4 Scout (MoE, 109B parameters with ~17B active) – exhibit larger errors, weaker pooled reliability, poorer calibration, and wider Bland–Altman limits, underscoring that effective capacity and the reasoning setup are critical for speech-level fidelity.

\end{document}